\documentclass[twoside,leqno,twocolumn]{article}

\usepackage[letterpaper]{geometry}

\usepackage{ltexpprt}
\usepackage{hyperref}

\usepackage{graphicx} 
\usepackage{natbib}  
\usepackage{caption}
\usepackage{verbatim}
\usepackage{algorithm}
\usepackage{algorithmic}
\usepackage{color}
\usepackage{textcomp}
\usepackage{xcolor}
\usepackage{subfigure}
\usepackage{gensymb}
\usepackage{mathtools, nccmath}
\usepackage{lipsum}
\usepackage{booktabs}  
\usepackage{placeins}
\usepackage{amsfonts,amsmath}
\usepackage{nicefrac}
\usepackage{paralist}
\begin{document}

\newcommand\relatedversion{}

\title{\Large Hub-VAE: Unsupervised Hub-based Regularization of Variational Autoencoders}
\author{Priya Mani\thanks{George Mason University, USA.}
\and Carlotta Domeniconi\thanks{George Mason University, USA.}}

\date{}

\maketitle

% Copyright Statement
% When submitting your final paper to a SIAM proceedings, it is requested that you include
% the appropriate copyright in the footer of the paper.  The copyright added should be
% consistent with the copyright selected on the copyright form submitted with the paper.
% Please note that "20XX" should be changed to the year of the meeting.

% Default Copyright Statement
\fancyfoot[R]{\scriptsize{Copyright \textcopyright\ 2023 by SIAM\\
Unauthorized reproduction of this article is prohibited}}

% Depending on which copyright you agree to when you sign the copyright form, the copyright
% can be changed to one of the following after commenting out the default copyright statement
% above.

%\fancyfoot[R]{\scriptsize{Copyright \textcopyright\ 20XX\\
%Copyright for this paper is retained by authors}}

%\fancyfoot[R]{\scriptsize{Copyright \textcopyright\ 20XX\\
%Copyright retained by principal author's organization}}

%\pagenumbering{arabic}
%\setcounter{page}{1}%Leave this line commented out.

\newcommand{\algcomment}[1]{{\color{blue}// #1}}
\newcommand{\hubs}{\mathcal{H}}

\newcommand{\lossContr}{\mathcal L_C}
\newcommand{\lossPerc}{\mathcal L_P}
\newcommand{\lossRec}{\mathcal L_R}
\newcommand{\lossKL}{\mathcal L_{\text{KL}}}
\newcommand{\mlpOut}{f^{(\tau)}}

\newcommand{\vc}{\mathbf c}
\newcommand{\vw}{\mathbf w}
\newcommand{\vx}{\mathbf x}
\newcommand{\vz}{\mathbf z}

\newcommand{\enc}{\mathbf Q^{(\phi)}}
\newcommand{\gen}{\mathbf P^{(\theta)}}
\newcommand{\hubenc}{\mathbf R^{(\phi)}}

\newcommand{\tree}{\mathcal T}
\newcommand{\ckmm}{C_{\text{CKMM}}}
\newcommand{\lca}{\text{LCA}}
\newcommand{\hcCoef}{C_{\text{HC}}}
\newcommand{\argmin}{argmin}

\newtheorem{definition}{Definition}
\newtheorem{remark}[lemma]{Remark}
\newtheorem{assumption}{Assumption}

\newcommand{\multiline}[1]{
        \begin{tabular}{@{}c@{}}
        #1
        \end{tabular}
}

\definecolor{darkyellow}{rgb}{1, 0.8, 0.0}

\begin{abstract} \small\baselineskip=9pt Exemplar-based methods rely on informative data points or prototypes to guide the optimization of learning algorithms. Such data facilitate interpretable model design and prediction. Of particular interest is the utility of exemplars in learning unsupervised deep representations. In this paper, we leverage hubs, which emerge as frequent neighbors in high-dimensional spaces, as exemplars to regularize a variational autoencoder and to learn a discriminative embedding for unsupervised down-stream tasks. We propose an unsupervised, data-driven regularization of the latent space with a mixture of hub-based priors and a hub-based contrastive loss. Experimental evaluation shows that our algorithm achieves superior cluster separability in the embedding space, and accurate data reconstruction and generation, compared to baselines and state-of-the-art techniques.
\end{abstract}

\section{Introduction}\label{sec:INTRO}
Traditional exemplar-based machine learning methods, such as the Parzen window estimator \cite{PARZEN}, nearest neighbor methods \cite{KNN}, and exemplar-SVMs \cite{EXEMPLAR-SVM}, rely on a set of exemplars and on a distance metric defined on the training data to learn a predictive model. Exemplars are instances selected from the training data which are informative for a given learning task. Exemplar-based methods improve the discriminative ability of a model, and support interpretable model design and prediction. In deep learning, exemplar-based methods are typically used to explain post-training the predictions of supervised black-box models (e.g., \cite{LIME}, \cite{DINO}). Such methods are model-agnostic, as the learned exemplars do not contribute to the training of a black-box model, but are instead used to make the model interpretable. A different approach in exemplar-based deep learning (the focus of this paper) is to leverage exemplars \textit{during} the training of a deep neural network to regularize the latent embedding (e.g., \cite{PROTONETS}, \cite{EXEMPLAR-VAE}, \cite{CLIQUE-CNN}). %A related line of research are attention-based methods (\cite{GAT}, \cite{SELF-ATTN}), which learns to weight specific instances or features in the data for a given learning task. 
%Most of the existing exemplar-based deep learning models are supervised. However, supervised models learn task-specific embeddings which are not applicable to different semantic tasks, and rely on training labels which are costly to obtain and prone to human annotation errors.  
In this paper, we leverage exemplars to train an unsupervised generative model, which learns discriminative latent embeddings suitable for unsupervised down-stream tasks.

%Auto-encoders \cite{AE} are the most popular deep learning framework for unsupervised models. An auto-encoder maps an input data point into a latent space representation through an encoder, and minimizes the error in reconstruction of the data point from its latent space representation through a decoder. A VAE learns a regularized latent embedding by modeling a prior distribution over the latent space. VAEs learns data distributions in the latent space instead of point estimates, which facilitates generative modeling using the decoder.
Variational autoencoders (VAEs, \cite{VAE}) are a popular deep learning framework for unsupervised generative models.
A VAE maps an input data point $\vx$ into a distribution $q_{\phi}(\vz|\vx)$ in the latent space (known as the \emph{variational posterior}), through an encoder network parameterized by $\phi$. 
The latent representation ${\bf{z}}$ is mapped into a distribution of the reconstructed input $p_{\theta}(\vx|\vz)$ through a decoder network parameterized by ${\theta}$.
A VAE regularizes the latent embedding by modeling a prior distribution ${p(\vz)}$ over the latent space.
%The distribution $q_{\phi}(\vz|\vx)$ is known as the variational posterior, as it approximates the intractable true conditional posterior through variational inference.%\todo{I don't understand the last part}
The objective when training a VAE is to maximize a lower bound on the log marginal likelihood of $\vx$, derived by variational inference:
% \begin{equation*}
% \label{eq:VAE}
% \begin{alignedat}{2}
% \mathcal{L(\phi,\theta; \vx)}= \mathbb{E}_{q_{\phi}(\vz|\vx)}[\log p_{\theta}(\vx|\vz)] - D_{KL}(q_{\phi}(\vz|\vx) || p(\vz))
% \end{alignedat}
% \end{equation*}
%\begin{equation*}
\[
    \label{eq:VAE}
    \mathcal{L(\phi,\theta; \vx)}= \mathbb{E}_{q_{\phi}(\vz|\vx)}[\log p_{\theta}(\vx|\vz)] - D_{{\text{KL}}}(q_{\phi}(\vz|\vx) || p_{\theta}(\vz)),
    \]
%\end{equation*}
where $D_{{\text{KL}}}(\cdot)$ is the KL-divergence between distributions. %{\color{red}{[If you don't reference this equation later in the paper, you can delete the number.]}}

%\todo{Some citations are broken}
Various approaches have been proposed in the literature to regularize the latent space of a VAE, e.g. learning a disentangled latent representation \cite{BETA-VAE}, \cite{FACTOR-VAE}, and learning an informative prior distribution \cite{EXEMPLAR-VAE}, \cite{VAMPPRIOR}.
The predictive performance of a model which uses latent embeddings for an unsupervised down-stream task depends on the separability of data clusters in the latent space. In this work we propose to use {\it {hubs}} as exemplars to drive the regularization of the latent space.
The hubness phenomenon \cite{HUBNESS} refers to the emergence of few data points that are frequent neighbors of the data. Such phenomenon affects nearest neighbor computations and the \textit{cluster assumption} \cite{CLUSTER-ASSUMPTION}.
%on data. Due to this, few data points emerge as \textit{hubs}, which are frequent $k$-nearest neighbors of the data.

To compute hubs in a collection of data, the concept of \emph{ hubness score} is used. The hubness score $N_k(\mathbf{x})$ of a data point $\mathbf{x}$ is defined as the number of points that have $\mathbf{x}$ as their $k$-nearest neighbor. The data points which contribute to the hubness of $\mathbf{x}$ are termed the \textit{reverse $k$-nearest neighbors} of $\mathbf{x}$. %\cite{HUBNESS}.}  %\gtodo{Explain what these are}.
%\end{definition}
A hub is a data point whose hubness score exceeds by a certain threshold the mean hubness score of the data:
%\begin{equation}
$N_k > \mu_{N_k} + \lambda \sigma_{N_k}$
%\end{equation}
where $\mu_{N_k}$ and $\sigma_{N_k}$ are the mean and standard deviation of the distribution of hubness scores within the data. Whether a hub has a positive or negative influence on data clustering depends on the degree of label mismatch between the hub and its reverse $k$-nearest neighbors.
\textit{Bad} hubs with greater than 50\% label mismatch appear near cluster boundaries and can cause the merging of unrelated clusters, while \textit{good} hubs are located near the true cluster centers, and can be useful seeds to guide data clustering \cite{ROLE-HUBS, HUBS-SUBSPACE}.
Previous work on regularization doesn't account for hubness in their objectives.
%[{\color{red} In Fig 1, keep only one caption and delete those associated to each subfig. CD}]

%\gtodo{Resize Fig. 1 so that legend is readable.}
\begin{figure*}[ht]
 \centering
    %  \subfigure[]%[TSNE plot of Hub-VAE (training at 10 epochs)]
    %  {
    %     \centering
    %     \includegraphics[width=35mm,height=30mm]{pics/tsnetrain_before_10epochs.png}
    %     \label{fig:example:clusters}
    % }
     \subfigure[]%[Chosen exemplars (Hub-VAE)]
     {
        \centering
        \includegraphics[width=29.5mm,height=30mm]{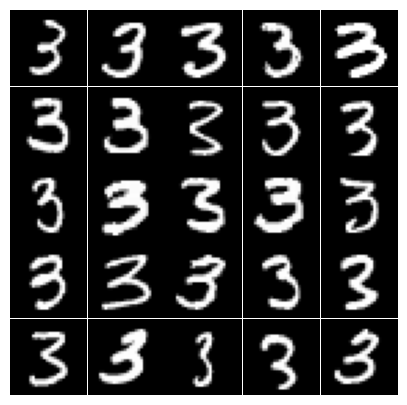}
        \label{fig:example:hub-vae}
    }
    \subfigure[]%[Chosen exemplars (Exemplar-VAE)]
    {
        \centering
        \includegraphics[width=29.5mm,height=30mm]{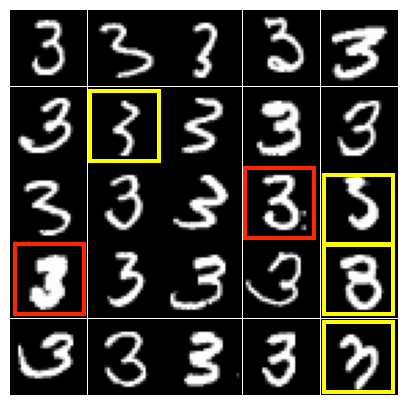}
        \label{fig:example:exemplar-vae}
    }
    \subfigure[]%[Chosen exemplars (Hub-VAE)]
    {
        \centering
        \includegraphics[width=35mm,height=30mm]{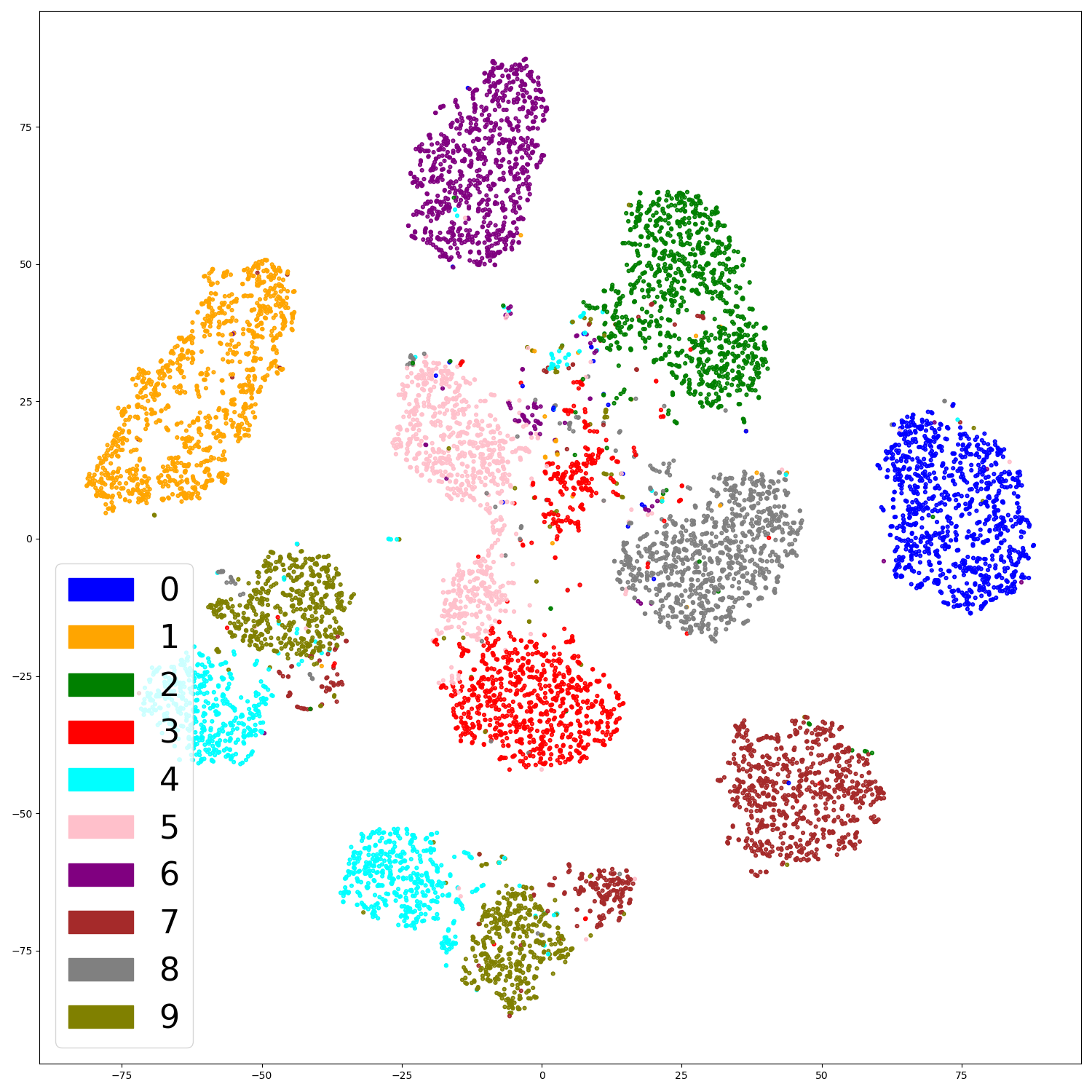}
        \label{fig:example:tsne-hubvae}
    }
    \subfigure[]%[Chosen exemplars (Exemplar-VAE)]
    {
        \centering
        \includegraphics[width=35mm,height=30mm]{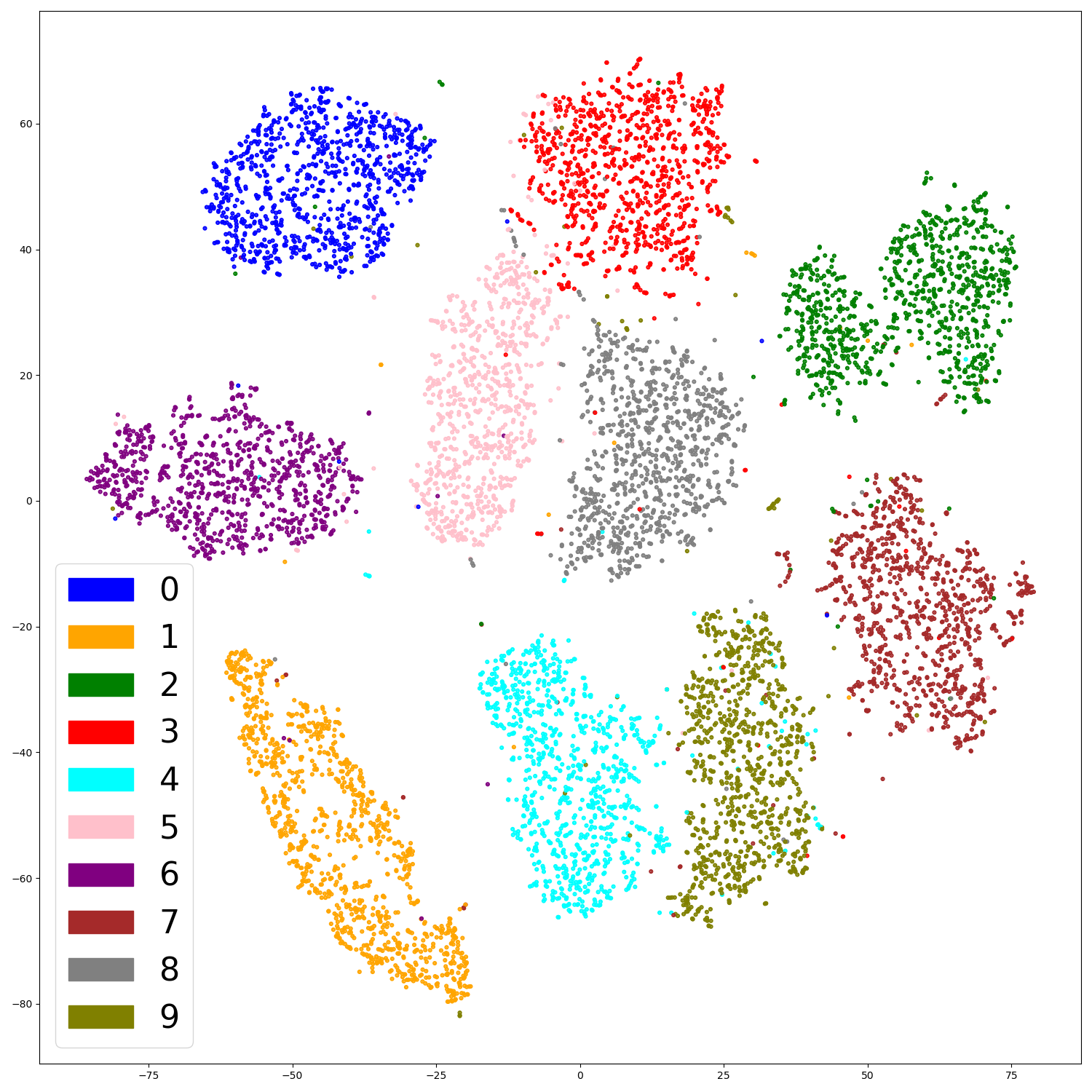}
        \label{fig:example:tsne-exemplar}
    }
    %  \subfigure[Sample of digit 3 exemplars learned by Hub-VAE]%[Chosen exemplars (Hub-VAE)]
    %  {
    %     \centering
    %     \includegraphics[width=29.5mm,height=30mm]{pics/hubs_3.png}
    % }
    % \subfigure[Sample of digit 3 exemplars learned by Exemplar-VAE]%[Chosen exemplars (Exemplar-VAE)]
    % {
    %     \centering
    %     \includegraphics[width=29.5mm,height=30mm]{pics/random_3.png}
    % }
    % \subfigure[TSNE plot of DMNIST test data embeddings learned by Hub-VAE]%[Chosen exemplars (Hub-VAE)]
    % {
    %     \centering
    %     \includegraphics[width=35mm,height=30mm]{pics/tsne_hubvae.png}
    % }
    % \subfigure[TSNE plot of DMNIST test data embeddings learned by Exemplar-VAE]%[Chosen exemplars (Exemplar-VAE)]
    % {
    %     \centering
    %     \includegraphics[width=35mm,height=30mm]{pics/tsneexm.png}
    % }
    %  \subfigure[TSNE plot of Hub-VAE (training at 100 epochs)]{
    % \centering
    % \includegraphics[width=30mm,height=30mm]{pics/tsnetrain_before_100.png}
    % }
    % \subfigure[TSNE plot of Exemplar-VAE (training at 100 epochs)]{
    % \centering
    % \includegraphics[width=30mm,height=30mm]{pics/tsneexmtrain_before_100.png}
    % }
    \caption{%(a) TSNE plot of DynamicMNIST training for Hub-VAE; 
    A sample of digit 3 exemplars learned by (a) Hub-VAE and (b) Exemplar-VAE. {\color{red}Red} bounding boxes indicate images with noise or defects and {\color{darkyellow}yellow} bounding boxes indicate images that have a large stroke variation. For example, one of the highlighted images can be mistaken as a member of class 8. (c)-(d) show the TSNE plots of DynamicMNIST test data embeddings learned by (c) Hub-VAE and (d) Exemplar-VAE. The TSNE plots are color-coded according to ground-truth. The clusters obtained by Hub-VAE are more compact and better separated, which lead to a superior clustering accuracy for digit 3: 87\% (Hub-VAE) vs. 83\% (Exemplar-VAE).
    %d)-(e) depicts the evolved latent embedding of training data at 100 epochs for Hub-VAE and Exemplar-VAE respectively.
    }
    % \caption{(a) TSNE plot of DynamicMNIST training for Hub-VAE; sample of digit 3 exemplars learned by (b) Hub-VAE, and (c) Exemplar-VAE. {\color{red}Red} bounding boxes indicate images with noise or defects in them, while {\color{darkyellow}yellow} bounding boxes indicate images that have large variation in their stroke. For example, one of the highlighted images could be confused as class 8.{\color{blue}(d)-(e) show the TSNE plot of DynamicMNIST test data embeddings learned by (d) Hub-VAE and (e) Exemplar-VAE. The color-code on TSNE plots denotes the ground-truth. }
    % %d)-(e) depicts the evolved latent embedding of training data at 100 epochs for Hub-VAE and Exemplar-VAE respectively.
    % }
    %\vspace{-1.0em}
    \label{fig:example}
\end{figure*}

In this paper, we regularize a VAE by leveraging good hubs as exemplars learned in the latent space.
We learn a discriminative latent embedding by modeling a mixture of hub-based priors, and an adversarial margin between bad neighbors (i.e. $k$-nearest neighbors with a label mismatch) via contrastive learning in the latent space \cite{CL}. 

\subsection{Design Considerations} Hubs exhibit characteristics which make them well-suited to be selected as exemplars to learn an  embedding space for clustering. In fact, good hubs facilitate the generation of prototypes of a given class, due to their proximity to cluster centers. A recent approach called Exemplar-VAE \cite{EXEMPLAR-VAE} uses exemplars chosen at random  to regularize a VAE by learning a mixture of exemplar-based priors distribution. 
%While a standard VAE only uses the decoder network in the generative process, Exemplar-VAE allows for conditional data generation through the exemplar-based prior. 
%Our proposed model (Hub-VAE) stands out in several ways. 
%The exemplars in \cite{EXEMPLAR-VAE} are chosen randomly from the training data; 
As such, the resulting components of the prior distribution may include outliers or data points at the boundaries of clusters, which may merge clusters in the learned embedding space. Furthermore, the chosen exemplars do not depend on the evolving latent space of the VAE. In contrast, our proposed approach computes hubs in the VAE latent space which evolves with training. 
%Hence our approach performs a data-driven selection of informative data points as exemplars. 
%Due to the useful clustering properties exhibited by good hubs, our proposed mixture of hub-based priors can capture the underlying clustering structure of the data. 
To further enhance the learned embeddings, Hub-VAE also optimizes a contrastive loss term that increases the margin between points and their nearest neighbors with a mismatch in  estimated labels. 

% As an example, consider the TSNE plot given in Fig.~\ref{fig:example}, showing the learned latent embedding of Dynamic MNIST data after training Hub-VAE.
% Digit 3 is similar to 2, 5 and 8, and can potentially be confused with digits from these classes in a down-stream clustering task.
Fig.~\ref{fig:example:hub-vae} shows a sample of hubs selected as exemplars by Hub-VAE from the class digit ``3" (Dynamic MNIST dataset). Similarly, Fig.~\ref{fig:example:exemplar-vae} shows a sample of exemplars chosen by Exemplar-VAE.
In Fig.~\ref{fig:example:exemplar-vae}, we highlight the images which could be confused as members of another class (e.g. class 8) and those which show a large variation from the prototypical image of digit 3.
The hubs chosen by Hub-VAE are closely aligned to prototypical shapes of digit 3, and have a better image quality, while the images chosen by Exemplar-VAE have a large stroke variability. 
In Fig.~\ref{fig:example:tsne-hubvae}-\ref{fig:example:tsne-exemplar}, we show the encoding of the test data for DynamicMNIST obtained by Hub-VAE and Exemplar-VAE, respectively. As a consequence of the chosen hubs as exemplars, the clusters learned by
Hub-VAE are more compact and  better separated. % and splits diverse images within a given class into separate clusters.
We computed the clustering accuracy of digit 3 on the encoded embedding of test data. We used $k$-means to cluster the data and assigned each cluster to the majority class label in the cluster.
 Exemplar-VAE results in an average accuracy of $0.83$ $(\pm 0.01)$ for digit class 3 and Hub-VAE gives an average accuracy of $0.87$ $(\pm 0.06)$. 
 % Hub-VAE achieves a significant improvement in digit accuracy compared to Exemplar-VAE. Exemplar-VAE merges images of digit 9 with clusters of other digits, while Hub-VAE groups the images of digit 9 into their own cluster.
 This experiment shows the potential of a hub-based regularization of VAEs, particularly for clustering tasks  which can benefit from the properties of hubs.
 
 %{\color{blue} Fig.~\ref{fig:example} (d)-(e) shows the latent embedding of training data at 100 epochs for Hub-VAE and Exemplar-VAE respectively. It can be observed that the latent clusters in the embedding of Hub-VAE are more globular than Exemplar-VAE. Hub-VAE splits the digit class 9 into two clusters to account for the variability in shape/stroke within its elements, while Exemplar-VAE learns a single non-globular cluster interspersed between other clusters. This explains the higher clustering quality of Hub-VAE on test data. Though Exemplar-VAE has a better separation between clusters, the learned embedding was not suitable for an unsupervised similarity based grouping which merged digit 9 into its nearby clusters for digits 4 and 7.  }
% \begin{table}[t]
% \caption{Average clustering accuracy for digit 9 of Dynamic MNIST. Standard deviations are given in paranthesis.}
% \begin{center}
% \begin{tabular}{c|c|c}
% %\hline
% %\textbf{Table}&\multicolumn{3}{|c|}{\textbf{Table Column Head}} \\
% %\cline{2-4} 
% %\textbf{Head} & \textbf{\textit{Table column subhead}}& \textbf{\textit{Subhead}}& \textbf{\textit{Subhead}} \\
% \textbf{Data} & \textbf{Exemplar-VAE} & \textbf{Hub-VAE}\\
% \hline
% Dynamic MNIST & 0.13 (0.10) & {\bf 0.55} (0.10)\\
% %\hline
% \end{tabular}
% \label{tab:example}
% \end{center}
% \end{table}

 %{\color{red}{[Are we showing the TSNE plot after 100 training epochs? Does it show a better separation among the clusters?]}}
\subsection{Contributions}
%\gtodo{Expand this section}
%\gtodo{Can we make this a bit more specific and quantitative in some of the places? E.g. we ran dataset X and got improvement Y\% or something along these lines?}

We present a novel approach (Hub-VAE) for the design of variational autoencoders, and leverage the hubness phenomenon to guide the regularization of the latent space via hub exemplars. This gives  an embedding space well-aligned with the clustering structure of the data, which also lead to accurate data reconstruction and generation.
Our main contributions are:
\begin{compactenum}
    \item \textbf{Unsupervised regularization and hub-based priors}: We propose an unsupervised and exemplar-based regularization framework for VAEs, built on a data-driven selection of hubs in the latent space. We use a mixture of hub-based priors to regularize the VAE latent space by selecting informative hubs as exemplars. 
    \item \textbf{Unsupervised contrastive loss}: We further regularize the VAE latent space via an adversarial margin between data points and neighbors with mismatched estimated labels, via a contrastive loss. %{\color{red}{[As earlier, this statement needs to be more precise.]}}
    \item \textbf{Improved cluster separability and generative modeling}: We empirically evaluate our method on data clustering, representation learning, and generative modeling. Our experiments show that Hub-VAE  is competitive against the state-of-the-art.
\end{compactenum}
%\dtodo{Add HC stuff here}
\section{Related Work} 
Variational auto-encoders are versatile, deep generative models with a wide range of applications \citep{VQ-VAE, GEN-FEW-SHOT, TD-VAE}. VAEs have a strong theoretical foundation and are easier to train and optimize than other classes of generative models \citep{GAN}. In this section we briefly discuss previous work on VAEs. 
\subsection{VAE Regularization}
%The regularization of the latent space of VAEs has been studied in the literature from several different perspectives. 
Several approaches have been proposed to learn a disentangled representation in the latent space and make the latent dimensions independent. The $\beta$-VAE \citep{BETA-VAE} weights the KL-divergence term by a factor $\beta$ to trade off learning an accurate reconstruction with disentanglement, %. \citep{TCBETA-VAE} isolates sources of disentanglement and introduces a total correlation loss into the VAE objective, 
while Factor-VAE \citep{FACTOR-VAE} encourages the representations to be factorial. 
Another way to learn an expressive and interpretable latent representation is to modify the prior distribution of the VAE. %Traditionally, a standard normal distribution is used as the prior. However, this leads to over-regularization of the latent space and results in poor generalization. 
\citep{Prior1, Prior2} have shown that the prior is essential to improving the performance of the VAE and several methods have proposed using a mixture of Gaussians prior. However, they do not allow for a closed-form optimization of the VAE objective. The authors in (VAE-Vampprior, \citep{VAMPPRIOR}) learn an approximation of the optimal mixture prior as a variational mixture of posteriors based on pseudo-inputs which are learned through the VAE optimization. The recent work in (Exemplar-VAE, \citep{EXEMPLAR-VAE}) learns a mixture of exemplar-based priors, where the exemplars are chosen from the training data, and proposes a framework for conditional image generation. The work in \citep{Cauchy} introduces a mixture prior with a Cauchy-Schwarz divergence term to regularize the VAE. The work (ByPE-VAE, \citep{ByPE-VAE}) conditioned the mixture prior on a Bayesian pseudo-coreset which is learned along with the optimization of VAE.
%\todo{Add ByPE-VAE}

\subsection{Contrastive Learning}
Contrastive learning is closely related to large margin nearest neighbors (LMNN, ~\citep{LMNN}) in traditional metric learning and optimizes a similar objective in a self-supervised or supervised deep-learning setting. A common approach is to select positive and negative samples with respect to an anchor data point, and to optimize the latent space so that positive samples are near each other in the learned representation, while negative samples are pushed farther away from the anchor. Several variants of contrastive losses have been proposed, such as to select positive samples by data augmentation \citep{CONTRASTIVE-AUG}, co-occurrence \citep{CONTRASTIVE-MULTIVIEW}, and to choose important triplets for learning \citep{TCUR}. \citep{RATIO} proposed an unsupervised deep metric learning method where a contrastive ratio between cluster centers is minimized.
\subsection{Hubness Phenomenon}
%\gtodo{maybe move to appendix?}
\cite{HUBNESS} provides a detailed theoretical study of the hubness phenomenon, its emergence, and its impact on learning tasks. Much of the previous work on hubness consider hubs as detrimental to their learning task (e.g., music retrieval \citep{HUBS-MUSIC}, finger-print identification \citep{HUBS-FINGERPRINT}, zero-shot learning \citep{HUBS-ZEROSHOT}), and aim to reduce hubness in data (\citep{HUB-REDUCTION}).
However, few approaches have attempted to leverage hubs for unsupervised learning. \citep{ROLE-HUBS} empirically analyzed the role of hubs w.r.t $k$-means clustering.
This work identified good hubs as cluster prototypes and designed several hub-based variants of $k$-means to leverage hubs as cluster centroids in $k$-means iterations.
\citet{HUBS-SUBSPACE} identified good hubs using a pre-trained classifier and leveraged them to optimize a subspace clustering algorithm.

\section{Hub-based VAE Regularization}
\begin{figure}[t]
%\vskip 0.2in
\begin{center}
\begin{small}
\centerline{\includegraphics[height=50mm,width=90mm]{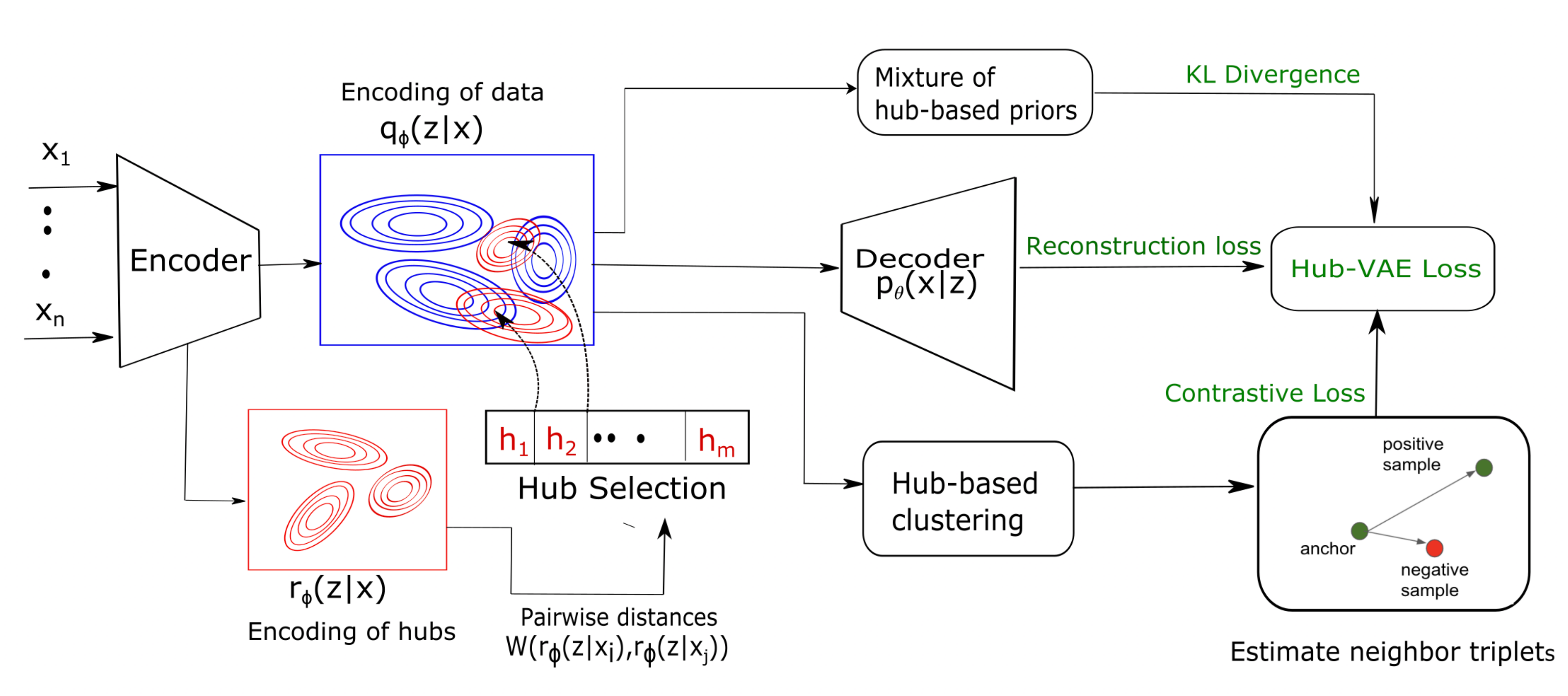}}
\end{small}
\end{center}
\vspace{-2em}
\caption{Hub-VAE architecture}
\label{fig:HUB-VAE}
\end{figure}
The architecture of Hub-VAE is given in Fig.~\ref{fig:HUB-VAE}. The model consists of an encoder  which maps data and hubs into their respective distributions in the latent space. The encoded data and hubs are reconstructed using a shared decoder.
%\gtodo{Cut MLP}
%An MLP network takes in input the pairwise distances between the two sets of distributions, corresponding to the data and hubs respectively, and learns the mixing coefficients of the hub-based priors. This design enables the MLP to learn an adaptive prior distribution for the given data.
%The MLP learns larger mixing coefficients for the hubs which are most relevant to learn a discriminative embedding, as well as an accurate reconstruction of the data.
The encoder and the decoder can be viewed as parametric functions which learn the parameters of the variational posterior and  hub-based prior, and the generative model respectively. 

We next formally define the components of Hub-VAE and how they are combined together. Let $\vx=\{\vx_i\}_{i=1}^{n}$ be the input data and ${\bf H}=\{\vx_h\}_{h=1}^{m}, m < n$ be the input representation of the hubs selected from the latent embedding of $\vx$, where $m$ is the number of components in the hub-based prior.  Each data point $\vx_i$ is encoded into a variational posterior distribution 
$q_{\phi}(\vz| \vx_i)=\mathcal{N}(\vz| \mu_{\phi}(\vx_i), \Sigma_{\phi}(\vx_i))$, while each hub $\vx_h$ is encoded into a component prior distribution 
$r_{\phi}(\vz| \vx_h)=\mathcal{N}(\vz| \mu_{\phi}(\vx_h), \sigma^2I)$ in the latent space, where $\mathcal{N(.)}$ denotes the Gaussian distribution, $\mu_{\phi}$ and $\Sigma_{\phi}$ denote parametric mean and covariance functions which are learned by the encoder network with parameter set $\phi$. %\gtodo{What is $\mu_\phi$ as a function?}
The sampled latent space variable is mapped back into the input space through a decoder using a generative model $p_{\theta}(\vx_i|\vz)$.
To simplify the presentation, we denote $\enc_i = q_{\phi}(\vz| \vx_i)$, $\hubenc_h = r_{\phi}(\vz| \vx_h)$ and $\gen_i = p_{\theta}(\vx_i|\vz)$.
%The specific form of the encoder distributions are given below:
%\begin{equation}
%\label{eq:FormalNotation}
%\begin{alignedat}{2}
%\text{Variational posterior}: q_{\phi}(\vz| \vx_i) %=\mathcal{N}(\vz| \mu_{\phi}(\vx_i), \Sigma_{\phi}({\bf %x}_i)) \\
%\text{Hub-based prior}: r_{\phi}(\vz| \vx_h) = %\mathcal{N}(\vz| \mu_{\phi}(\vx_h), \sigma^2I)\\
%\text{Generative model}: p_{\theta}(\vx_i|\vz)
%\end{alignedat}
%\end{equation}
%where $\mathcal{N(.)}$ denotes a Gaussian distribution. 
The encoder networks share the same parameter set $\phi$. The prior distribution is defined as an isotropic Gaussian with scalar variance $\sigma^2$, while the variational posterior has a diagonal covariance $\Sigma$. The mean and variance of the variational posterior, as well as the mean of the mixture of hub-based priors are learned.  The specific form of the distribution of the decoder depends on the type of data; typically a Bernoulli distribution is used for binary or discrete data and a Gaussian distribution is used for continuous data. 
\S \ref{HUB-SELECT}, \S \ref{HUB-PRIOR} and \S \ref{HUB-MARGIN} describe each component of Hub-VAE in detail. \S \ref{HUB-OBJECTIVE} describes the overall objective function of Hub-VAE.  The pseudocode for Hub-VAE is given in Algorithm 1 in the supplementary material.

\subsection{Hub Selection}
\label{HUB-SELECT}
We compute the hubs at the beginning of each training epoch, and use them to learn a mixture prior distribution. Hubs are computed within each mini-batch of an epoch from the $k$-nearest neighbor graph ($k$-NN) of the data, as defined in \S\ref{sec:INTRO}. A variational encoder learns data distributions instead of point estimates in the latent space. Hence, we construct the $k$-NN graph based on the 2-Wasserstein \cite{2-WASSERSTEIN} distance between the learned distributions in the latent space. Given two distributions $p_i({\bf \mu}_i,{\bf \Sigma}_i)$ and $p_j({\bf \mu}_j,{\bf \Sigma}_j)$ with means ${\bf \mu}_i$, ${\bf \mu}_j$, and diagonal covariances ${\bf \Sigma}_i$, ${\bf \Sigma}_j$ respectively, the 2-Wasserstein distance between the distributions is defined as 
%follows:
%\begin{equation}
%\label{eq:2-Wasserstein}
%\begin{alignedat}{2}
$W(p_i, p_j) = (||{\bf \mu}_i - {\bf \mu}_j||_2^2 + ||{\bf \Sigma}_i^{\nicefrac{1}{2}} - {\bf \Sigma}_j^{\nicefrac{1}{2}}||_F^2)^{\nicefrac{1}{2}}$
%\end{alignedat}
%\end{equation}
where $||.||_F$ denotes the Frobenius norm.
%\begin{definition}
We set $\lambda=0.5$ and $k=\sqrt{B}$ to compute the hubness scores $N_k$ (defined in \S\ref{sec:INTRO}), where $B$ is the number of data points in a mini-batch.
% \begin{definition}
%     \emph{The bad hubness score} of a data point is computed as the percentage of its reverse $k$-nearest neighbors  with a label mismatch \cite{HUBNESS}.  %\gtodo{Explain what these are}.
% \end{definition}
%\begin{definition}
   
%\end{definition}

As discussed in \S \ref{sec:INTRO}, not all hubs exhibit useful clustering properties. While \textit{good} hubs appear near the centers of the true clusters within data, a \textit{bad} hub has a significant label mismatch with its reverse $k$-nearest neighbors (R$k$NN), and can negatively affect tasks involving similarity or nearest neighbor computations. \emph{A bad hub} is a hub whose label mismatches with more than 50\% of its reverse k-nearest neighbors.% \cite{HUBNESS}.  %\gtodo{Explain what these are}. 
Identifying bad hubs without access to class labels is a challenge. 
%{\color{blue} There is little previous work on unsupervised characterization of good and bad hubs. The work in \cite{HUBS-SUBSPACE} learned a pre-trained classifier to identify good and bad hubs using meta-features of hubs. However, their method was applied as a pre-processing step to distinguish between hubs in the input space.}  
We address this problem by formulating a scoring function to filter bad hubs based on the characteristics exhibited by hubs in the latent space. Good hubs tend to be located near dense regions; hence the pairwise distances between the distributions of good hubs and their reverse nearest neighbors tend to be smaller then those of bad hubs. As a consequence, the average probability of reconstructing a good hub from the distribution of its reverse $k$-nearest neighbors is higher than for bad hubs. Based on this, we formulate a scoring function for good hubs: %Eq.~\eqref{eq:HUB-SCORING}.
\begin{equation}
\label{eq:HUB-SCORING}
\begin{alignedat}{2}
G(\vx_h) = \frac{\sum_{r\in \text{RkNN}(\vz_h)} p_{\theta}(\vx_h|\vz_r)}{\sum_{r\in \text{RkNN}(\vz_h)} W(q_{\phi}(\vz|\vx_r), q_{\phi}(\vz|\vx_h))}, \nonumber
\end{alignedat}
\end{equation}
% \begin{equation}
% \begin{alignedat}{2}
% B(\vx_h) = \left(\frac{\sum_{r\in \text{RkNN}(\vz_h)} p_{\theta}(\vx_h|\vz_r)}{N}\right)^{-1} +\\ 
% \frac{\sum_{r\in \text{RkNN}(\vz_h)} W(q_{\phi}(\vz|\vx_r), q_{\phi}(\vz|\vx_h))}{N}
% \end{alignedat}
% \end{equation}

%\gtodo{Not clear why this is a good scoring function. What is the motivation behind the choices and some other options here? How is this normalized, etc?}
\noindent where $q_{\phi}$ and $p_{\theta}$ denote the distributions learned by the encoder and decoder respectively; $\vx_h$ and $\vx_r$  denote the input representations of a hub and its reverse $k$-nearest neighbor; $\vz_h$ and $\vz_r$ denote the data points sampled in the latent space from their encoded distributions.% {\color{blue} Min-max normalization is applied to the numerator and denominator of Equation.~\eqref{eq:HUB-SCORING} so that their values are in the range [0, 1].} 
%Our hub selection is unsupervised and hence can be applied within the optimization of the auto-encoder to identify good and bad hubs in the evolving latent space.
%\gtodo{Can we increase the size of Figure 3?}
\begin{figure*}[t]
 \centering
      \subfigure[Dynamic MNIST]{
        \centering
        \includegraphics[width=40mm,height=30mm]{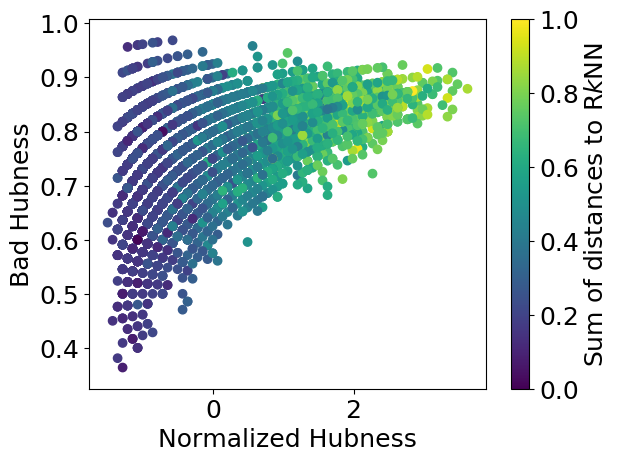}
        }
        % \subfigure[USPS]{
        % \centering
        % \includegraphics[width=40mm,height=30mm]{pics/USPS_sumdis_new.png}
        % }
        %  \subfigure[Caltech101Silhouettes]{
        % \centering
        % \includegraphics[width=40mm,height=30mm]{pics/CALTECH_sumdis_new.png}
        % }\\
        \subfigure[Fashion MNIST]{
        \centering
        \includegraphics[width=40mm,height=30mm]{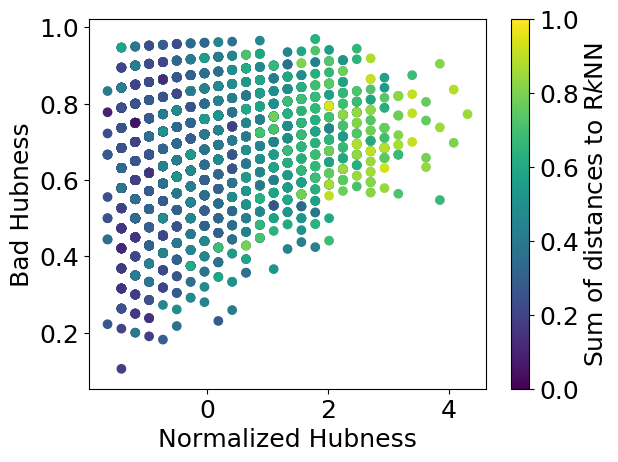}
        }
        \subfigure[Dynamic MNIST]{
        \centering
        \includegraphics[width=40mm,height=30mm]{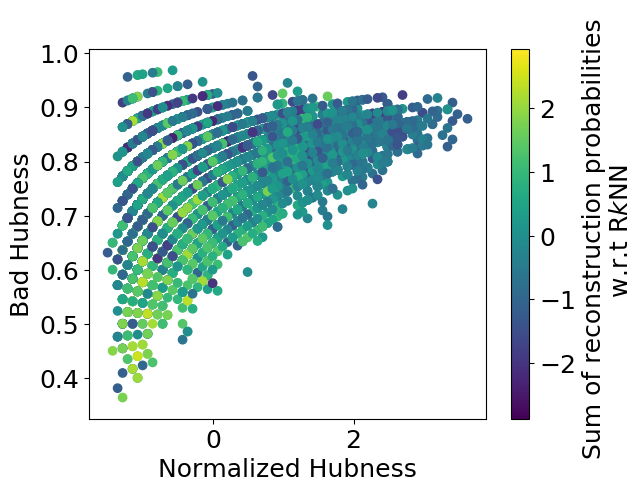}
        }
      \subfigure[Fashion MNIST]{
        \centering
        \includegraphics[width=40mm,height=30mm]{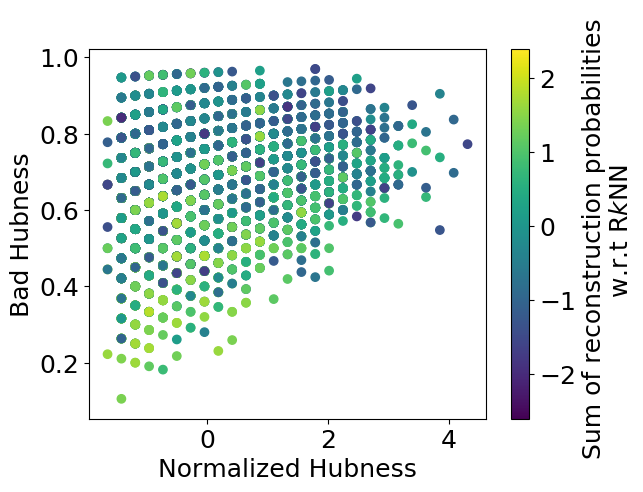}
        }
        % \subfigure[USPS]{
        % \centering
        % \includegraphics[width=40mm,height=30mm]{pics/USPS_reconprob_new.png}
        % }
        %  \subfigure[Caltech101Silhouettes]{
        % \centering
        % \includegraphics[width=40mm,height=30mm]{pics/CALTECH_reconprob_new.png}
        %}
        
        \caption{ Scatter plots of characteristics of hubs.
        %pairwise 2-Wasserstein distances between hubs and their R$k$NNs in the latent space; (e)-(h) Scatter plots of reconstruction probabilities of hubs with respect to the distributions of their R$k$NNs.
        The $x$-axis in each sub-plot denotes normalized ($\mu_{N_k}$ = 0, $\sigma_{N_k}$ = 1) hubness scores. The $y$-axis denotes bad hubness.
        The hubs are color-coded by the sum of the pairwise distances to their reverse $k$-nearest neighbors (R$k$NN) in plots (a)-(b), and by their reconstruction probabilities w.r.t. the distributions of their R$k$NN in plots (c)-(d).
        Hubs with high pairwise distances (top-right quadrant of (a)-(b)) and low reconstruction probabilities  (bottom-left quadrant of (c)-(d)) with respect to their R$k$NN are strong bad hubs. %The distinction of strong bad hubs through the measured characteristics is more pronounced for DynamicMNIST, FashionMNIST, and Caltech101Silhouettes.
        % {\color{red}{Need to add labels of axes and fix current naming. Need to clarify how "bad hubness" is computed in this}}
        }
        %\vspace{-1.0em}
    \label{fig:HUB-SCORING}
\end{figure*}
%\gtodo{What is ``bad hubness'' in Fig 3 caption?}
%\gtodo{For figures we need to provide interpretation of results.}
 
To motivate the choice of the scoring function above, consider the plots given in Figure~\ref{fig:HUB-SCORING} which show the hubs computed in an epoch and their characteristics for the datasets used in our experiments. 
The $x$-axis in each sub-plot of Figure~\ref{fig:HUB-SCORING} denotes normalized ($\mu_{N_k}$ = 0, $\sigma_{N_k}$ = 1) hubness scores. The $y$-axis denotes bad hubness. The bad hubness score of a hub is computed as the percentage of its R$k$NN with a label mismatch. For the purpose of this analysis, we use the ground truth labels to compute bad hubness.
% \dtodo{I commented the line before
        The hubs are color-coded by the sum of the pairwise distances to their R$k$NN in plots (a)-(d), and by their reconstruction probabilities w.r.t. the distributions of their R$k$NN in plots (e)-(h).
 %In each sub-plot of Figure~\ref{fig:HUB-SCORING}, we plot the hubs selected in an epoch with the standardized hubness score \gtodo{What is this, same as the above formula?} on the $x$-axis and the bad hubness score on the $y$-axis. 
 %{\color{blue} The bad hubness score of a hub is computed as the percentage of its R$k$NN with a label mismatch.}  
  %$\frac{\sum_{\vx_i \in RkNN(\vx_h)} label(\vx_i) \neq label(\vx_h)}{}$} 
 We can see that the hubs with high pairwise distances  (top-right quadrant of (a)-(d)) and low reconstruction probabilities  (bottom-left quadrant of (e)-(h)) with respect to their reverse $k$-nearest neighbors are strong bad hubs.% {\color{blue}The distinction of strong bad hubs through the measured characteristics is more pronounced for DynamicMNIST, FashionMNIST, and Caltech101Silhouettes.}
 %\gtodo{Where and how do we look to see this?}. 
 The above characteristics exhibited by hubs in the latent space of VAE, motivate the coice of the scoring function $G(\vx_h)$ above. We use a threshold to select good hubs based on $G(\vx_h)$, setting the threshold adaptively to $\frac{\text{max}(z\text{-score}(G))}{2}$. %We identify such bad hubs as those that give a low score value of the function in Eq. (\ref{eq:HUB-SCORING}).

Prior to each training epoch, we pre-compute a pool of good hubs by selecting them from each mini-batch in the epoch. The data in each mini-batch of an epoch are encoded into the latent space using the configuration of the VAE model trained in the previous epoch. The training of the VAE in the current epoch is then carried out, and uses the selected pool of hubs to regularize the latent space.  %Note that the hub computation is performed as a step prior to training the VAE for the current epoch.  

\subsection{Hub-based Prior Distribution}\label{HUB-PRIOR}
We learn a mixture of \textit{good} hub-based priors to regularize the VAE. Each mini-batch in a training epoch samples a subset of hubs from the selected pool of hubs for that epoch. %The clustering property of good hubs enables learning an informative prior distribution which facilitates the VAE to learn a discriminative latent embedding where the data clusters are better separated. 
We define the component prior distribution corresponding to each hub as an isotropic Gaussian with a mean function which depends on the chosen hub and shared covariance among all of the hub distributions, as also done in previous work on exemplar based priors~\cite{EXEMPLAR-VAE, VAMPPRIOR}:
%\begin{equation}
%\label{eq:HUB-PRIOR}
%\begin{alignedat}{2}
$\hubenc_{h} = r_{\phi}(\vz| \vx_h) = \mathcal{N}(\vz| \mu_{\phi}(\vx_h), \sigma^2I)$,
%\end{alignedat}
%\end{equation}
where $\sigma$ is a scalar.
Based on these distributions, we define the KL-divergence loss between the variational posterior and the hub-based prior as 
%\begin{equation}
    $\lossKL(\vx_i, \phi) = D_{{\text{KL}}}(\enc_i || \frac{1}{m}\sum_{j=1}^{m} \hubenc_{h_j})$.
%\end{equation}
%The input and output of the MLP for a given data point $\vx_i$, and the final mixing coefficients computed from $f_{\tau}(\vx_i)_j$, are given below.
%\begin{equation}
%\label{eq:MLP}
%\begin{alignedat}{2}
%MLP: n-RelU-\frac{n}{2}-RelU-n-Softmax\\
%Input: W(q_{\phi}(\vz|\vx_i), r_{\phi}(\vz|\vx_h_j)), ~j=1,2, \dots m\\
%Output: f_{\tau}(\vx_i)_j, j=1,2, \dots m,  ~~s.t. \sum_{j=1}^{m} f_{\tau}(\vx_i)_j=1 \\
%Mixing~coefficients: w_j = \frac{c_j}{\sum_j {c_j}}, j=1,2, \dots m,  \\c_j = \sum_{i=1}^{B} f_{\tau}(\vx_i)_j 
%\end{alignedat}
%\end{equation}
%where $f_{\tau}(.)$ represents the output of MLP as a function with parameter $\tau$, and $B$ denotes the total number of data points in a mini-batch.
The reconstruction loss component of the VAE objective
$
\lossRec(\vx_i, \phi, \theta) = -\mathbb{E}_{\enc_i}[\log \gen_i]
$
further regularizes the latent space. The complete loss function for Hub-VAE is given in \S~\ref{HUB-OBJECTIVE}. 
In Hub-VAE a hub is chosen from the selected pool of hubs in the latent space, and is encoded into a latent representation using the hub-based prior. The decoder then reconstructs the latent representation as a new sample. For generative modeling, Hub-VAE requires the decoder network as well as the learned pool of hubs and the hub-based prior $r_{\phi}$.

\subsection{Hub-based Contrastive Learning}\label{HUB-MARGIN}
We regularize the VAE to learn a discriminative latent space embedding by increasing the margin between a point and its neighbor when a label mismatch is detected. This regularization is  useful in high-dimensional latent spaces which are affected by the emergence of bad hubs. We achieve this through contrastive learning within the $k$-neighborhood of each point. Contrastive learning works similarly to the hinge loss in \cite{LMNN}. Positive and negative samples are selected w.r.t an anchor point, depending on the label match/mismatch between the anchor and the samples. The distances between each triplet (anchor, positive, and negative sample) are adjusted such that the distance of the anchor to the negative sample is larger than that of the positive sample by a certain margin. 

Contrastive learning is typically applied in a supervised setting where the class labels of data points are known. Our model is unsupervised which brings forth the challenging problem of selecting good and bad neighbors without access to their labels. To overcome this issue, we cluster the data in the latent space and use the clustering labels as a proxy for the ground truth. Since good hubs appear near cluster centers, we seed the clustering process using the hub sample to obtain a higher clustering quality.
In each mini-batch, the hubs are clustered using $k$-means\footnote{Number of clusters in $k$-means is set to the number of classes.}, and the rest of the data are assigned to the cluster label of its nearest hub, as computed by the 2-Wasserstein distance between their distributions. 

%\dtodo{Make it clear that we need to know the number of classes}
Given the cluster labels, we choose triplets from the neighborhood of each data point as follows: each data point is an anchor, the farthest neighbor that shares  the same cluster label with the anchor is chosen as the positive sample, and the neighbors with label mismatch are chosen as the negative samples. The contrastive loss of a given anchor point $\vx_a$ with $t$ triplets $(\vx_a, \vx_p, \vx_{n_i}), i=1,2, \dots t$ is defined as: 
\begin{equation}
\label{eq:CONTRASTIVE}
\begin{alignedat}{2}
\lossContr(\vx_a, \phi) &= \sum_{i=1}^{t} [W(\enc_a, \enc_p) - W(\enc_a, \enc_{n_i}) + 1]_{+} \\
&\{ \vx_p, \vx_{n_i}\} \in \text{$k$NN}(\vx_a), i=1,2,...\dots t\\
&l(\vx_a) \neq l(\vx_{n_i}), l(\vx_a) = l(\vx_p) \nonumber
\end{alignedat}
\end{equation}
%\dtodo{What if there is no $\vx_p$ or $\vx_{n_i}$ among the nearest neighbors?}
%\dtodo{Clarify how we select $\vx_p$ (closest one) and $\vx_{n_i}$ (all the negatives)}
where $\vx_p$ is the positive sample, $\vx_{n_i}$ are negative samples associated with the anchor, and $l(\vx)$ is the label of $\vx$. 
The loss is minimized when the distance from the anchor to each negative example is greater then the distance to the positive example by $1$.
Hence, training using this loss makes the anchor (i.e. its embedding) significantly closer to the positive example than to negative examples\footnote{Since we use cluster labels to choose $\vx_p$ and $\vx_{n_i}$, each anchor $\vx_a$ will have a positive sample within its neighborhood. When $\{ \vx_{n_i}\}$ is an empty set,  $\lossContr(\vx_a, \phi) = 0$.}.

A training mini-batch is a sample of data points, and the distribution of cluster densities in the sample could vary.
Hence we compute adaptive neighbors for each data point to reduce over-stepping cluster boundaries due to the selection of $k$ nearest neighbors.
We select triplets from an adaptive neighborhood of each data point in the latent space.
The adaptive neighbors are computed as follows: we construct a $k$-nearest neighbor graph and compute mean $\mu_D$ and standard deviation $\sigma_D$ of the distribution of $k^{th}$-neighbor distances across the data points, and eliminate the neighbors whose distances exceed $\mu_D + 2 \sigma_D$.
The hub seeded clustering together with the adaptive neighborhood facilitates selection of more accurate triplets for contrastive learning.

\subsection{Hub-VAE Objective Function}\label{HUB-OBJECTIVE}
The objective function minimized by Hub-VAE is given below. It includes the following terms: (1) the reconstruction loss between the data point given in input to the encoder and the output of the decoder; (2) the KL-divergence between the variational posterior and the hub-based prior; (3) the contrastive loss within the neighborhood of the input data point. %The overall loss function is obtained by summing the component losses over each data point: 
% \begin{equation}
% \label{eq:HUB-VAE}
% \begin{alignedat}{2}
%     L_{\text{HUB-VAE}}(\phi, \theta; \vx_a) =  \lossRec(\vx_a, \phi, \theta)   + \lossContr(\vx_a, \phi) + \beta \lossKL(\vx_a, \phi) \nonumber
% \end{alignedat}
% \end{equation}
\begin{align*}
    &L_{\text{HUB-VAE}}(\vx_a; \phi, \theta) =\\
    &\lossRec(\vx_a, \phi, \theta)   + \lossContr(\vx_a, \phi) + \beta \lossKL(\vx_a, \phi)
    % \label{eq:HUB-VAE}
\end{align*}
%The loss function in Eq. (\label{eq:HUB-VAE}) is equivalent to the negative of the marginal likelihood of $\vx_a$ with additional loss terms for negative entropy and contrastive loss. 
The training of Hub-VAE is conducted in mini-batches, and each mini-batch minimizes $L_{\text{HUB-VAE}}$ over the data points in that mini-batch. The reparameterization~\cite{VAE} is applied to $q_{\phi}(\vz|\vx_a)$ to enable back-propagation through the encoder.
Hub-VAE performs stochastic optimization using the Adam algorithm \cite{ADAM} with normalized gradients \cite{ADAM-NORM}.
Training is performed for 100 epochs and a validation loss is computed at the end of each epoch using the current model configuration and a separate validation data.
Similarly to previous models proposed in the literature \cite{EXEMPLAR-VAE, VAMPPRIOR}, the validation loss uses all of the learned hubs as exemplars and does not incorporate the contrastive loss term in its objective%
\footnote{$\lossContr$ isn't included in the validation objective as its computation requires knowledge of the number of classes in the data. Furthermore, $\lossContr$ acts as an additional regularizer, and the learning of VAE can be approximated by $L_{\text{val}}$, without sacrificing model performance.} %and hence $L_{\text{HUB-VAE-VAL}}$ can be used an approximation for VAE learning.}:
\begin{equation}
\label{eq:HUB-VAE-VAL}
\begin{alignedat}{2}
L_{\text{val}}(\vx_v; \phi, \theta) =  \lossRec(\vx_v, \phi) + \beta D_{{\text{{KL}}}}(\enc_v || \frac{1}{|\hubs|} \sum_{j=1}^{|\hubs|} \hubenc_j)\nonumber
\end{alignedat}
\end{equation}
%\gtodo{Can we elaborate on why is it ok to have training and validation losses to be different?}
%\dtodo{Also, is it OK that the weights are $\frac 1m$ instead of $\mlpOut(\vx_v)$?}
where $\hubs$ is the pool of selected good hubs and $\vx^v$ is a validation instance. The configuration which results in the minimum validation loss is used as the final model.

\section{Experiments}\label{sec:Exp}

\begin{table*}[ht!]
    \begin{center}
    \begin{tabular}{c|c|c|c|c|c}
    %\hline
    \textbf{Data} & \textbf{VAE-Gaussian} & \textbf{VAE-Vamp} & \textbf{Ex-VAE} & \textbf{ByPE-VAE} & \textbf{Hub-VAE}\\
    %& \textbf{Gaussian} & \textbf{Vampprior} &  &   &\\
    \hline
    DMNIST & 0.52 (0.02) & 0.58 (0.01) & 0.64 (0.01) & 0.66 (0.01) & {\bf \underline{0.73}} (0.02)\\
    USPS  & 0.52 (0.01) & 0.64 (0.01) & 0.69 (0.01) & 0.68 (0.01) & {\bf \underline{0.76}} (0.01)\\
    FMNIST & 0.57 (0.01) & 0.58 (0.01) & 0.59 (0.01) & 0.61 (0.01) & {\bf \underline {0.64}} (0.01) \\
    Caltech101 & 0.59 (0.002) & 0.60 (0.002) & 0.60 (0.002) & {\bf \underline{0.61}} (0.001) &  0.60 (0.004)\\
    %\hline
    %AVG & 0.57 & 0.59 & 0.62 &  & {\bf 0.68}\\
    %\hline
    \end{tabular}
    \end{center}
    \caption{$k$-means V-measure. We show mean (std) over 10 runs. Statistically significant results are underlined.}
    \label{tab:V-MEASURE}
\end{table*}
\begin{figure*}[ht!]
 \centering
      \subfigure[Hub-VAE ]{
        \centering
        \includegraphics[width=25mm,height=25mm]{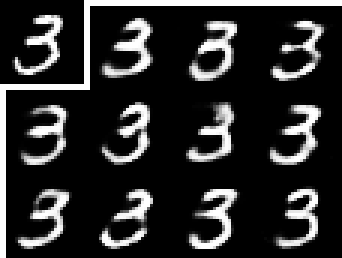}
        }
    %   \subfigure[Hub-VAE]{
    %     \centering
    %     \includegraphics[width=30mm,height=30mm]{pics/generated_4_hub2.png}
    %     }
    \subfigure[Exemplar-VAE ]{
        \centering
        \includegraphics[width=25mm,height=25mm]{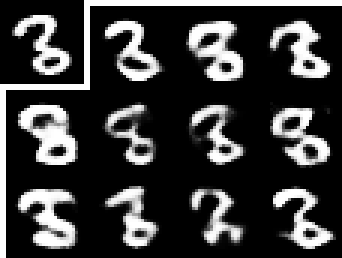}
        }
        \subfigure[ByPE-VAE ]{
        \centering
        \includegraphics[width=25mm,height=25mm]{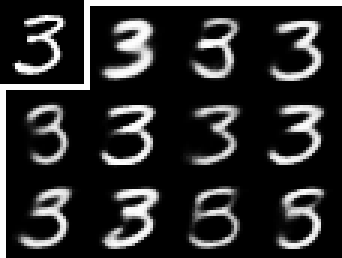}
        }
        \subfigure[Hub-VAE]{
        \centering
        \includegraphics[width=25mm,height=25mm]{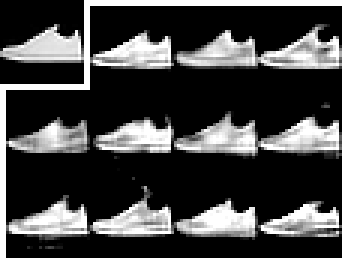}
        }
        % \subfigure[Hub-VAE]{
        % \centering
        % \includegraphics[width=30mm,height=30mm]{pics/generated_9_hub2.png}
        % }\\
    %   \subfigure[Exemplar-VAE]{
    %     \centering
    %     \includegraphics[width=30mm,height=30mm]{pics/generated_4_exm1.png}
    %     }
        \subfigure[Exemplar-VAE]{
        \centering
        \includegraphics[width=25mm,height=25mm]{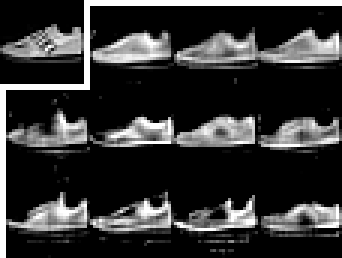}
        }
        % \subfigure[Exemplar-VAE]{
        % \centering
        % \includegraphics[width=30mm,height=30mm]{pics/generated_9_exm1.png}
        % }
        \subfigure[ByPE-VAE]{
        \centering
        \includegraphics[width=25mm,height=25mm]{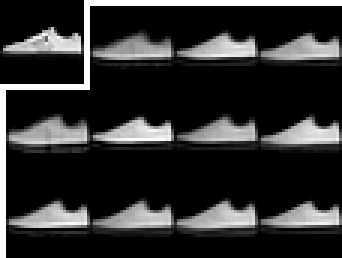}
        }
        \caption{Conditional image generation from classes $3$ in DMNIST and 'Sneaker' in FMNIST. The inset shows reference images.}
        %\vspace{-1.0em}
    \label{fig:GENERATION1}
\end{figure*}
We compare the methods on four well-known benchmark datasets: Dynamic MNIST (DMNIST)~\cite{DMNIST}, Fashion MNIST (FMNIST)~\cite{FMNIST}, USPS\footnote{\url{https://pytorch.org/vision/stable/datasets.html}}, and Caltech101Silhouettes\footnote{\url{https://people.cs.umass.edu/~marlin/ data/caltech101_silhouettes_28_split1.mat}.} (Caltech101). %Unlike \cite{EXEMPLAR-VAE}, we do not apply dynamic binarization on Fashion MNIST and use the gray-scale images as input.
%Table~\ref{tab:data} gives the summary statistics for the data.
We evaluate Hub-VAE against a baseline standard-VAE with a factored Gaussian distribution which is compared in \cite{VAMPPRIOR}, and state-of-the-art VAE regularization methods which modify the prior distribution: ByPE-VAE \cite{ByPE-VAE}, Exemplar-VAE \cite{EXEMPLAR-VAE}, denoted as Ex-VAE, and VAE-Vampprior \cite{VAMPPRIOR}. 
%e also designed a modification of Exemplar-VAE  with contrastive learning added to its objective, called Exemplar-VAE-Contrastive. We compute the clustering labels for triplet selection in Exemplar-VAE-Contrastive by standard $k$-means clustering. We designed this variant to assess the role played by hubs in VAE regularization. 
For all experiments, we set the number of components $m=1000$ in the prior distribution, and the latent space dimensionality $d=40$ \footnote{Number of coresets in ByPE-VAE, and the number of pseudo-inputs in VAE-Vampprior are set to 1000 for fair comparison.}. All methods are run for a maximum of 100 epochs. We use early-stopping with a look-ahead of 50 epochs. The initial value of $\beta$ in $L_{\text{HUB-VAE}}$ is set to 1 and is annealed as the epochs progress. Results are shown on average over $10$ runs and statistical significant results are underlined. All experiments are run on a A100 GPU on the Google Cloud Platform. Hub-VAE takes on average 28 sec to run an epoch for DMNIST, 37 sec for FMNIST, and 5 sec for USPS and Caltech101. Dataset summary and comparison of running times across all methods are given in the supplementary material.

\subsection{Representation Learning}
We evaluate the efficacy of the learned latent embeddings for unsupervised tasks using the V-measure \cite{V-MEASURE} and KNN purity. The V-measure  is an entropy-based evaluation metric computed as the harmonic mean of homogeneity and completeness score of a given clustering. Homogeneity measures the degree to which the members of a cluster belong to the same class, while the completeness score measures the degree to which the members of a class belong to the same cluster. %We use the implementation of V-measure in scikit-learn library\footnote{\url{https://scikit-learn.org/stable/modules/generated/sklearn.metrics.v_measure_score.html}} for our evaluation. 
KNN purity measures  the \% of $k$-nearest neighbors which share the same class label, i.e. the degree to which the cluster assumption holds on a given test data.  
The number of classes for $k$-means clustering is set equal to the true number of classes in the data. %We conducted 30 iterations of $k$-means on each run to compute an average V-measure for each training run, which are then averaged over 10 runs.  

%{\color{blue}We compare the results against state-of-the-art VAE methods which use a factored prior distribution.  
Tables~\ref{tab:V-MEASURE} show the results on V-measure. The results for KNN purity are given in the supplementary material. Hub-VAE outperforms baselines by at least 8\% for V-measure on DMNIST, FMNIST, and USPS, and attains comparable performance on Caltech101. Hub-VAE further achieves superior performance in V-Measure compared to Ex-VAE and ByPE-VAE for the large majority of the datasets, thus showing the advantage of using hubs as exemplars to improve cluster separability in the embedding space. %$t$-SNE plots of the encoded test data of DMNIST for Hub-VAE and Exemplar-VAE are shown in Fig.~\ref{fig:example}(c)-(d). 
We provide an ablation study, $t$-SNE plots, and additional results on CIFAR-10 data in the supplementary material.
% Among the competitors, Exemplar-VAE achieves the highest V-measure. In Fig.~\ref{fig:TSNE}, we compare the latent embedding of test data learned by Exemplar-VAE and Hub-VAE for Dynamic MNIST, Fashion MNIST, and USPS. The plot for Caltech101Silhouettes is not shown due to avoid clutter due to the presence of 101 classes in the data. We observe that the clusters learned by Hub-VAE are more compact and better separated than Exemplar-VAE, which is also reflected in the KMeans results.

% \begin{figure*}[t]
%  \centering
%       \subfigure[Dynamic MNIST (Hub-VAE)]{
%         \centering
%         \includegraphics[width=40mm,height=30mm]{pics/DMNIST_Hub_Tsne.png}
%         }
%       \subfigure[Fashion MNIST (Hub-VAE)]{
%         \centering
%         \includegraphics[width=40mm,height=30mm]{}
%         }
%         \subfigure[USPS (Hub-VAE)]{
%         \centering
%         \includegraphics[width=40mm,height=30mm]{pics/USPS_Exm_Hub.png}
%         }
        
%         \\
%         \subfigure[Dynamic MNIST (Exemplar-VAE)]{
%         \centering
%         \includegraphics[width=40mm,height=30mm]{pics/DMNIST_Exm_Tsne.png}
%         }
%       \subfigure[Fashion MNIST (Exemplar-VAE)]{
%         \centering
%         \includegraphics[width=40mm,height=30mm]{pics/FMNIST_Exm_Tsne.png}
%         }
%         \subfigure[USPS (Exemplar-VAE)]{
%         \centering
%         \includegraphics[width=40mm,height=30mm]{pics/USPS_Exm_TSNE.png}
%         }
%         \caption{TSNE plots of the latent embedding of test data learned by: (a)-(c) Hub-VAE, (d)-(f) Exemplar-VAE. }
%         \vspace{-1.0em}
%     \label{fig:TSNE}
% \end{figure*}
\subsection{Reconstruction and Generation}
%In this section, we qualitatively analyze the effectiveness of the distribution learned by the VAE decoder network. 
%Note that the reconstruction of digit 3 is incorrect for all the competitors, whose reconstructions resemble the digit 8. 
In Fig.~\ref{fig:GENERATION1}, we compare the quality of exemplar-based generative modeling using conditional image generation on Hub-VAE, Exemplar-VAE, and ByPE-VAE. We depict generated images from class 3 for DMNIST and from class 'Sneaker' for FMNIST. We observe that some of the images generated by Exemplar-VAE and ByPE-VAE resemble digits 8 or 5. These classes exhibit features which can confound their members, and they are embedded in close proximity in the latent space learned by Exemplar-VAE and ByPE-VAE.
In contrast, Hub-VAE results in accurate generations and closely follow the shape of the reference image (shown in the inset). Hub-VAE separates digits with varied strokes within a class into separate clusters, which helps obtain more accurate reconstructions and generations. Similarly, Exemplar-VAE does not fully capture the shape of 'Sneaker' and also generates blurred images. 
%The images generated by Hub-VAE are less blurred and closely follow the shape of the reference image (shown in the inset).
%Exemplar-VAE often generates blurred images and sometimes incorrect images (e.g., images resembling digit 8 are generated from digit 3).
The above results show that a hub-based regularization of VAEs is useful to improve the quality of the generative modeling. 
%Similarly, we plot the classes 'Sandals', 'Bag', 'Ankle Boots' and 'Shirt'. The classes 'Sandals' and 'Ankle Boots' can be confused with each other. We again observe the superior generative quality of Hub-VAE compared to Exemplar-VAE which generates blurred images and does not accurately capture the fine details of the reference image. 
% The generative ability of Exemplar-VAE may improve with a larger number of components in the prior distribution: \citet{EXEMPLAR-VAE} use a model with $\ge$ 10000 components for DMNIST and FMNIST.
% However, this also shows the strength of Hub-VAE which achieves high-quality results with a smaller number of components.

%We compare image reconstructions of methods in the supplementary material.
We compute Fretchet Inception Distance (FID) \cite{FID} between original and reconstructed images to evaluate their quality. In Table~\ref{tab:fid}, we observe that Hub-VAE has higher reconstruction quality than its competitors for DMNIST and USPS and similar quality for FMNIST.
\begin{table}[t]
    \begin{center}
    \begin{tabular}{c|c|c|c}
    %\hline
    \textbf{Data} &  \textbf{Hub-VAE} & \textbf{Ex-VAE} & \textbf{ByPE-VAE}\\
    %& \textbf{Gaussian} & \textbf{Vampprior} &  &   &\\
    \hline
    DMNIST & {\bf{\underline{8.45}}} (0.2) & 9.49 (0.6) & 21.39 (0.6) \\
    USPS  & {\bf{\underline{16.44}}} (0.3) & 20.12 (11) & 37.80 (2.6) \\
    FMNIST & 25.25 (0.2) & {\bf{25.15}} (0.3) & 54.90 (0.6) \\
    Caltech101 & 91.45 (0.6) & {\bf{\underline{87.01}}} (2.2) & 124 (3.1)\\
    %\hline
    %AVG & 0.57 & 0.59 & 0.62 &  & {\bf 0.68}\\
    %\hline
    \end{tabular}
    \end{center}
    \caption{FID on reconstructed images. Lower values signify higher quality. Statistically significant results are underlined.}
    \label{tab:fid}
\end{table}

\section{Sensitivity Analysis} 
\begin{figure}[t]
 \centering
        \includegraphics[width=65mm, height=50mm]{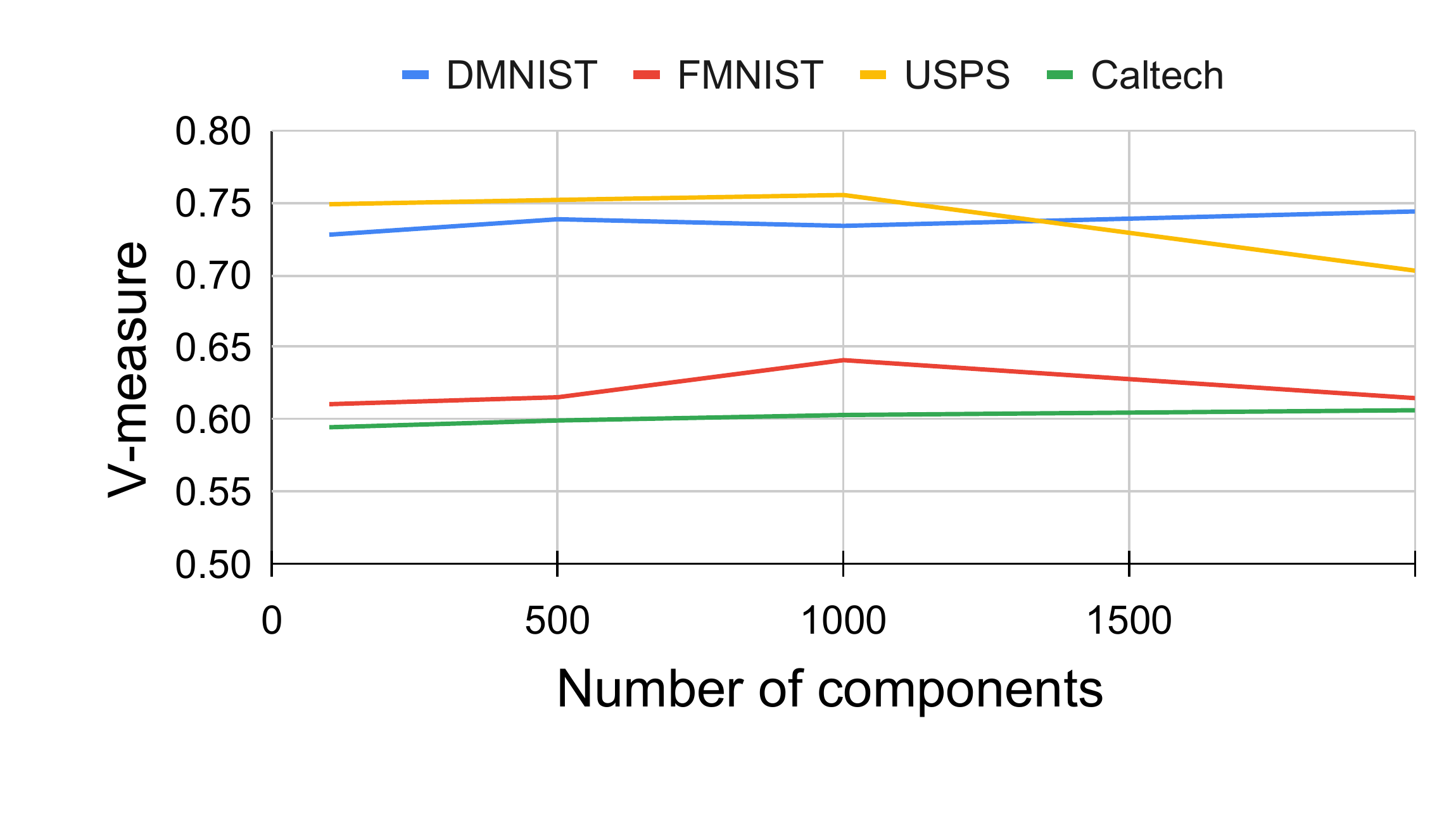}
        \caption{ Sensitivity analysis on number of components.}
        \label{fig:sensitivity_analysis}
        \end{figure}

We conduct sensitivity analysis to assess the influence of the number of components of the mixture prior distribution on VAE regularization. Figure~\ref{fig:sensitivity_analysis} shows the values of $k$-means V-measure across different number of components $m \in \{100, 500, 1000, 2000\}$. While we see a notable increase in V-measure for FMNIST when $m$ is increased from $500$ to $1000$, the trend is  stable for the other datasets for this range of $m$ values.
 %\vspace{-1em}       
\section{Conclusion}
We proposed a data-driven, hub-based regularization approach for VAEs to learn a discriminative and interpretable latent embedding using a mixture of hub based priors and leveraging the useful clustering properties of hubs.
We further enhanced the regularization by introducing a hub-based contrastive loss to increase the margin between bad neighbors in the latent space. 
Our method achieves competitive results against state-of-the-art VAE regularization methods.

\bibliographystyle{sdm23}
\bibliography{sdm23}
\end{document}

\newcommand{\algcomment}[1]{{\color{blue}// #1}}
\newcommand{\hubs}{\mathcal{H}}

\newcommand{\lossContr}{\mathcal L_C}
\newcommand{\lossPerc}{\mathcal L_P}
\newcommand{\lossRec}{\mathcal L_R}
\newcommand{\lossKL}{\mathcal L_{\text{KL}}}
\newcommand{\mlpOut}{f^{(\tau)}}

\newcommand{\vc}{\mathbf c}
\newcommand{\vw}{\mathbf w}
\newcommand{\vx}{\mathbf x}
\newcommand{\vz}{\mathbf z}

\newcommand{\enc}{\mathbf Q^{(\phi)}}
\newcommand{\gen}{\mathbf P^{(\theta)}}
\newcommand{\hubenc}{\mathbf R^{(\phi)}}

\newcommand{\tree}{\mathcal T}
\newcommand{\ckmm}{C_{\text{CKMM}}}
\newcommand{\lca}{\text{LCA}}
\newcommand{\hcCoef}{C_{\text{HC}}}
\newcommand{\argmin}{argmin}

\newtheorem{definition}{Definition}
\newtheorem{remark}[lemma]{Remark}
\newtheorem{assumption}{Assumption}

\newcommand{\multiline}[1]{
        \begin{tabular}{@{}c@{}}
        #1
        \end{tabular}
}

\definecolor{darkyellow}{rgb}{1, 0.8, 0.0}
\newcommand\relatedversion{}

\title{\Large Hub-VAE: Unsupervised Hub-based Regularization of Variational Autoencoders\\ Supplementary Material}
\author{Priya Mani\thanks{George Mason University, USA.}
\and Carlotta Domeniconi\thanks{George Mason University, USA.}}

\date{}

\maketitle

% Copyright Statement
% When submitting your final paper to a SIAM proceedings, it is requested that you include
% the appropriate copyright in the footer of the paper.  The copyright added should be
% consistent with the copyright selected on the copyright form submitted with the paper.
% Please note that "20XX" should be changed to the year of the meeting.

% Default Copyright Statement
\fancyfoot[R]{\scriptsize{Copyright \textcopyright\ 2023 by SIAM\\
Unauthorized reproduction of this article is prohibited}}

\section{Algorithm}
\label{app:algorithm}
Algorithm~\ref{app:algorithm} shows the pseudo-code of training epoch for Hub-VAE.
\begin{algorithm}
\caption{Hub-VAE}
\label{pseudocode}
% \begin{footnotesize}
\begin{algorithmic}[1]
    %\Procedure{H-WAMS}{}
    \STATE {\bfseries input:} Data $\vx=\{\vx_i\}_{i=1}^{n}$, number of components in the prior $m$, \# epochs =100,  mini-batch size $B=100$, $\beta=1$, variance of prior distribution $\sigma^2$, the number of clusters $K$, learned hubs $\hubs$
    \STATE {\bfseries output:} Learned parameters $\phi$, $\theta$, $\tau$ \\
    \STATE Initialize parameters $\phi$, $\theta$, $\tau$
    \FOR{ each training epoch }
    %\STATE \COMMENT{Hub Selection}\\
        \FOR{ each mini-batch $\{\vx_i\}_{i=1}^{B}$ in the epoch}
            \STATE Compute distributions $\hubenc_i$ and $\gen_i$ for each $\vx_i$ in the batch
            \STATE \algcomment{Compute pairwise 2-Wasserstein distances for all points in the batch}
            \STATE $d_{ij} \gets W(\hubenc_i, \hubenc_j), i, j \in [B]$
            \STATE \algcomment{Compute hubs}
            \STATE $\hubs \gets \{\vx_h \mid N_{\sqrt B}(\vx_h) > \mu + 0.5 \sigma\} $
            \STATE \algcomment{Estimate good hubness score}\\
            $G(\vx_h) \gets \frac{\sum_{r\in \text{RkNN}(\vz_h)} p_{\theta}(\vx_h|\vz_r)}{\sum_{r\in \text{RkNN}(\vz_h)} d_{rh}}$ for $h \in \hubs$
        \ENDFOR
        \STATE Eliminate bad hubs $\vx_{h_j}$ from the pool $\{\vx_{h}\}$:\\
        $\hubs \gets \{h \in \hubs \mid z\text{-score}(G(\vx_{h})) < \frac{\text{max}(z\text{-score}(G))}{2} \}$\\
        \FOR{ each mini-batch $\{\vx_i\}_{i=1}^{B}$ in epoch}
            \STATE Compute distribution $\enc_i$ for each $\vx_i$ in the batch
            \STATE Let $h_1, \ldots, h_m$ be a uniformly random sample from $\hubs$ \\
            %\STATE Let $\{h_j\}_{j=1}^{m}$ be a uniformly random sample of hubs from $\hubs$
            \STATE Let $l(\vx_{h_j})$ be the label after using $K$-Means clustering on the hubs $\vx_{h_1}, \ldots, \vx_{h_m}$, where $K$ denotes the number of clusters in the data
            \STATE \algcomment{Assign clustering labels $\{l_i\}$ to the closest hub}\\
            For each $i\in [B]$, let $l(\vx_i) \gets l(\vx_{h_j})$, where $j \gets \argmin_{j} \tilde d_{i h_j}$, where $\tilde d_{i h_j} = W(\enc_i, \hubenc_{h_j})$ \\
            %\dtodo{What is the distance here?}
            \STATE For each $a \in [B]$, compute the contrastive loss $\lossContr(\vx_a, \phi)$ %according to Eq.~\eqref{eq:CONTRASTIVE}
            \STATE Compute distribution $\gen_a$ for all $a \in [B]$\\
            \STATE Compute the loss $L_{\text{HUB-VAE}}(\vx_a; \phi, \theta, \tau)$
            \STATE Back-propagate and update parameters $(\phi, \theta, \tau)$ to minimize $L_{\text{HUB-VAE}}(\vx_a; \phi, \theta, \tau)$
        \ENDFOR 
    \ENDFOR
    \end{algorithmic}
% \end{footnotesize}
\end{algorithm}

\section{Additional Experiments}
\paragraph{Data}
Table~\ref{tab:data} gives the summary statistics for the data. We evaluated our method on an additional dataset, CIFAR-10 \footnote{https://www.cs.toronto.edu/~kriz/cifar.html}. 
\begin{table*}[ht!]
\begin{center}
\begin{tabular}{c|c|c|c}
%\hline
%\textbf{Table}&\multicolumn{3}{|c|}{\textbf{Table Column Head}} \\
%\cline{2-4} 
%\textbf{Head} & \textbf{\textit{Table column subhead}}& \textbf{\textit{Subhead}}& \textbf{\textit{Subhead}} \\
\textbf{Data} & \textbf{\# instances} & \textbf{\# input features} & \textbf{\# classes}\\
\hline
DMNIST & 70000 & 784 & 10\\
FMNIST & 70000 & 784 & 10\\
USPS  & 9298 & 256 & 10\\
Caltech101 & 8671 & 784 & 101 \\
CIFAR-10 & 60000 & 3072 & 10 \\
%\hline
\end{tabular}
\end{center}
\caption{Dataset information}
\label{tab:data}
\end{table*}

\subsection{Representation Learning}
We evaluate $k$-NN purity of data in Table~\ref{tab:KNN-PURITY}. The value of $k$ for computing the KNN purity is set to $k=\sqrt{n_{\text{test}}}$, where $n_{\text{test}}$ is the number of instances in the test data. Hub-VAE outperforms baselines by at least 1\% for KNN purity on DMNIST, FMNIST, and USPS, and attains comparable performance on Caltech101. 
\begin{table*}[ht!]
    \begin{center}
    \begin{tabular}{c|c|c|c|c|c}
    %\hline
    \textbf{Data} & \textbf{VAE-Gaussian} & \textbf{VAE-Vamp} & \textbf{Ex-VAE} &  \textbf{ByPE-VAE} & \textbf{Hub-VAE}\\
    %& \textbf{Gaussian} & \textbf{Vampprior} &  &   &\\
    \hline
    DMNIST & 72.77 (0.83) & 86.18 (0.12) & 88.24 (0.44) & {\bf{90.09}} (0.42)  & 89.36 (0.97)\\
    USPS  & 71.17 (0.68) & 80.94 (0.28) & 84.02 (0.59)  & 81.33 (0.19) &  {\bf \underline{85.79}} (0.57)\\
    FMNIST & 69.71 (0.19) & 71.55 (0.06) & 71.69 (0.10) & 71.37 (0.22) & {\bf \underline{72.92}} (0.15)\\
    Caltech101 & 41.80 (0.06) & 41.89 (0.08) & 42.16 (0.07) & {\bf{\underline{42.52}}} (0.08)  & 42.20 (0.11)\\
    %\hline
    %AVG & 65.68 & 69.73 & 70.93 &  & {\bf 72.45}\\
    %\hline
    \end{tabular}
    \end{center}
    \caption{KNN Purity. We show mean (std) over 10 runs. Statistically significant results are underlined.}
    \label{tab:KNN-PURITY}
\end{table*}

\begin{figure}[t]
    \centering
    \includegraphics[scale=0.2]{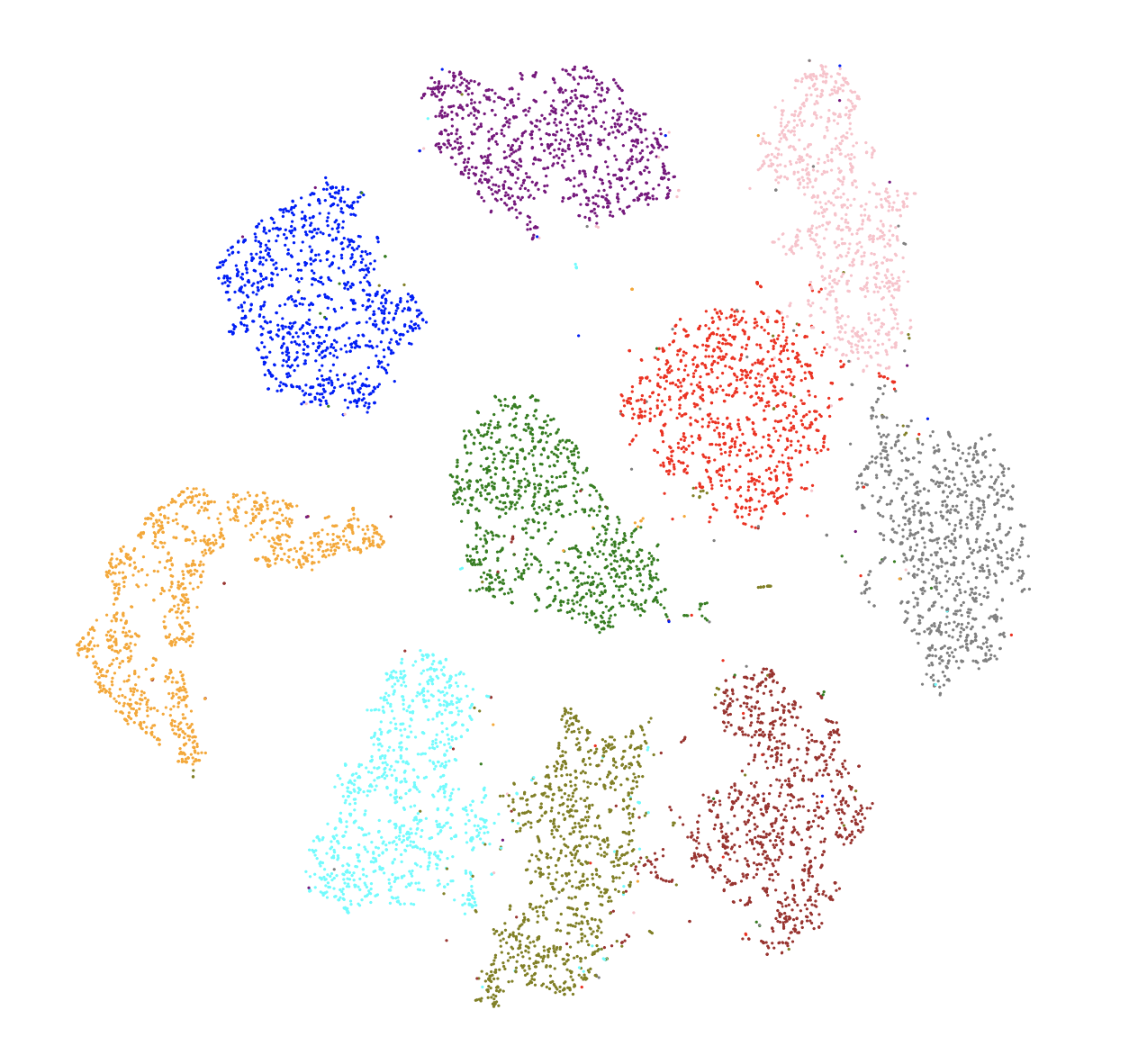}
    \caption{$t$-SNE plot of DMNIST test data for ByPE-VAE.}
    \label{fig:tsne_bypevae}
\end{figure}

Fig.~\ref{fig:tsne_bypevae} shows the $t$-SNE plot of test data of DMNIST for ByPE-VAE. Similar to Exemplar-VAE, ByPE-VAE does not form compact clusters.

In Table~\ref{tab:cifar10}, we evaluate $k$-means V-measure and KNN purity for CIFAR-10 across 5 independent training runs. We trained each method for 500 epochs for CIFAR-10. Hub-VAE outperforms the baseline and has nearly similar performance to ByPE-VAE. 
\begin{table*}[ht!]
\begin{center}
\begin{tabular}{c|c|c|c|c|c}
%\hline
%\textbf{Table}&\multicolumn{3}{|c|}{\textbf{Table Column Head}} \\
%\cline{2-4} 
%\textbf{Head} & \textbf{\textit{Table column subhead}}& \textbf{\textit{Subhead}}& \textbf{\textit{Subhead}} \\
\textbf{Metric} & \textbf{VAE-Gaussian} & \textbf{VAE-Vamp} & \textbf{Ex-VAE} & \textbf{ByPE-VAE} & \textbf{Hub-VAE}\\
\hline
$k$-means V-measure & 0.11 (0.002) & 0.10 (0.001) & 0.10 (0.004) & {\bf \underline{ 0.12} }(0.004)  &  0.11 (0.001) \\
KNN Purity & 21.65 (0.07) & 22.15 (0.05)  & 21.9 (0.06) & {\bf \underline{22.71}} (0.16) & 22.05 (0.04)   \\
%\hline
\end{tabular}
\end{center}
\caption{Evaluation of representation quality of CIFAR-10 embeddings. We show mean (std) over 5 runs. Statistically significant results are underlined.}
\label{tab:cifar10}
\end{table*}

\subsection{Ablation Study}

We evaluate the influence of different components of Hub-VAE on its objective function. We compute $k$-means V-measure and KNN purity for our model without applying contrastive loss (Hub-VAE-NoContrastive), and without applying hub selection (Hub-VAE-NoSelection). Table~\ref{tab:Ablation-Kmeans} and Table~\ref{tab:Ablation-KNN}%\gtodo{Can we highlight statistically significant results?}
%For Dynamic MNIST seems like -NoContrastive is doing better? Also, can we do the -NoMLP only run?} 
 show the $k$-means V-measure and KNN purity for these variants. 
 
 USPS and FMNIST show a significant decrease in performance for HuB-VAE variants. The performance drop is more pronounced for Hub-VAE-NoContrastive, which suggests that contrastive learning is useful in learning the underlying clustering structure of data by separating bad neighbors in the data.  Hub-VAE-NoSelection results in lower V-measure and KNN purity than Hub-VAE, which shows that use of good hubs as exemplars in the mixture prior plays a significant role in learning the clustering structure of data. DMNIST and Caltech101 did not show much change in performance on the variants. A lack of decrease in performance of Hub-NoSelection compared to Hub-VAE indicates that DMNIST and Caltech101 have fewer strong bad hubs in the data which can negatively influence the neighbor computation and regularization. The lack of performance drop without contrastive learning (Hub-VAE-NoContrastive) further indicates that these datasets have relatively fewer bad neighbors than FMNIST and USPS.  

%We also compare against the variant Exemplar-VAE-Contrastive, which has a contrastive loss term added to the loss function of Exemplar-VAE, as described in Section~\ref{sec:Exp}. Hub-VAE outperforms Exemplar-VAE-Contrastive, which shows that the mixture of hub-based priors also plays an important role in regularizing the VAE.}

%Hub-VAE-NoContrastive-NoMLP has a lower V-measure and KNN purity on average than Hub-VAE-NoContrastive, which shows that an adaptively weighted prior distribution is useful to improve the latent space quality. The normalization of the mixing coefficients over $m=1000$ components may have lowered their effect on the prior distribution, which may explain the smaller difference between the results of Hub-VAE-NoContrastive and  Hub-VAE-NoContrastive-NoMLP. %{[\color{red}{To make the comparison easier, you want to include the results of Hub-VAE in these tables as well.}]}

\begin{table*}[ht!]
\begin{center}
\begin{tabular}{c|c|c|c}
%\hline
\textbf{Data} &\textbf{Hub-VAE} & \textbf{Hub-VAE} &  \textbf{Hub-VAE} \\
& \textbf{-NoSelection} & \textbf{-NoContrastive} &  \\
%& & \textbf{-NoMLP} & \\
\hline
DMNIST &  0.73 (0.02) &  0.73 (0.002)  & 0.73 (0.02)\\
USPS  & 0.73 (0.01)  & 0.68 (0.01)  & {\bf \underline{0.76}} (0.01) \\
FMNIST & 0.62 (0.01) & 0.60 (0.005)  & {\bf \underline{0.64}} (0.01)\\
Caltech101 & 0.60 (0.001) &  0.60 (0.004) &  0.60 (0.004)\\
%\hline
%AVG & 0.64 & 0.65 & {\bf 0.68}

%\hline
\end{tabular}
\end{center}
\caption{ $k$-means V-measure on Hub-VAE variants. We show mean (std) over 10 runs. Statistically significant results are underlined.}
\label{tab:Ablation-Kmeans}
\end{table*}

\begin{table*}[ht!]
    \begin{center}
    \begin{tabular}{c|c|c|c}
    %\hline
    \textbf{Data} &\textbf{Hub-VAE} & \textbf{Hub-VAE} &  \textbf{Hub-VAE} \\
& \textbf{-NoSelection} & \textbf{-NoContrastive} &  \\
   % & & \textbf{-NoMLP} &\\
    \hline
    DMNIST & 89.07 (0.02) & {\bf \underline {90.36}} (0.002) &   89.36 (0.97)\\
    USPS  & 84.17 (0.46) &  83.92 (0.45) & {\bf \underline{85.79}} (0.57)\\
    FMNIST &  72.30 (0.14) & 71.83 (0.0)  & {\bf \underline{72.92}} (0.15)\\
    Caltech101 & 41.97 (0.03) & {\bf \underline {42.28}} (0.0) & 42.20 (0.11)\\
  %  \hline
   % AVG & 71.44 & 71.57 &  {\bf 72.45}
    \end{tabular}
    \end{center}
    \caption{ KNN purity on Hub-VAE variants. We show mean (std) over 10 runs. Statistically significant results are underlined.}
    \label{tab:Ablation-KNN}
\end{table*}

\subsection{Reconstruction and Generation}
\begin{figure*}[ht!]
 \centering
    \subfigure[Original Image ]{
        \centering
        \includegraphics[width=30mm,height=30mm]{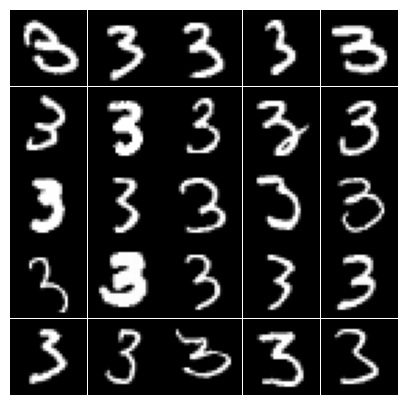}
        }
      \subfigure[Hub-VAE ]{
        \centering
        \includegraphics[width=30mm,height=30mm]{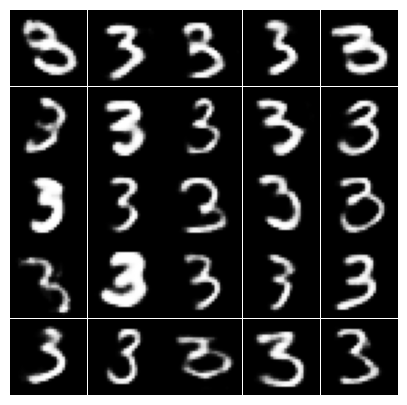}
        }
      \subfigure[Exemplar-VAE]{
        \centering
        \includegraphics[width=30mm,height=30mm]{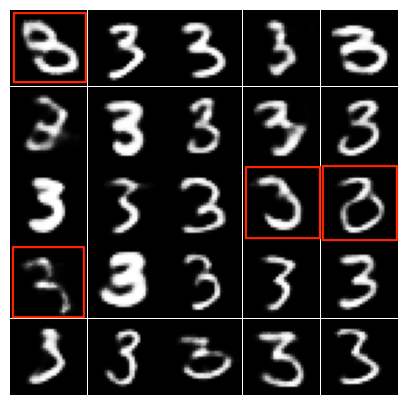}
        }
        \subfigure[ByPE-VAE]{
        \centering
        \includegraphics[width=30mm,height=30mm]{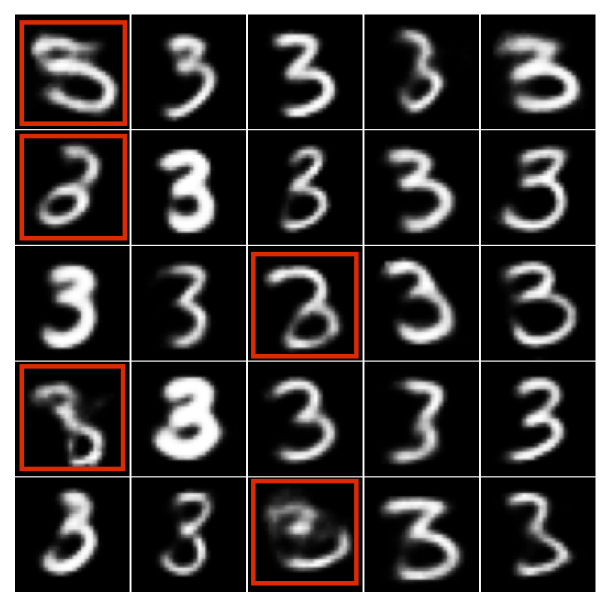}
        } \\
        \subfigure[Original Image ]{
        \includegraphics[width=30mm,height=30mm]{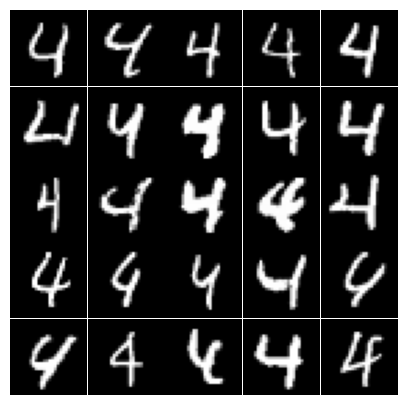}
        }
      \subfigure[Hub-VAE ]{
        \centering
        \includegraphics[width=30mm,height=30mm]{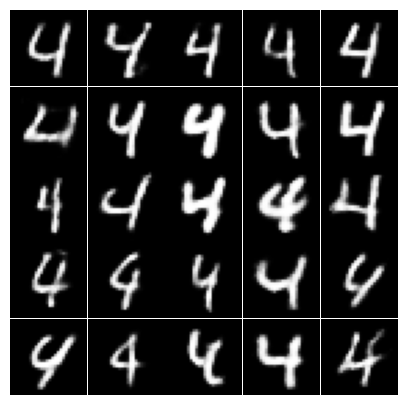}
        }
      \subfigure[Exemplar-VAE]{
        \centering
        \includegraphics[width=30mm,height=30mm]{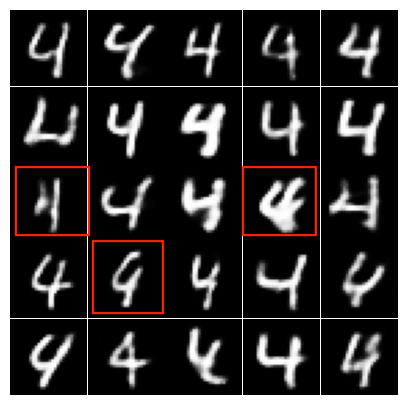}
        }
        \subfigure[ByPE-VAE]{
        \centering
        \includegraphics[width=30mm,height=30mm]{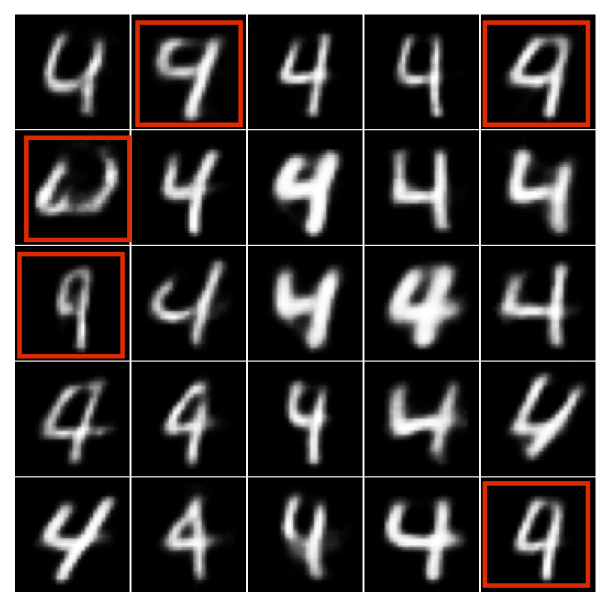}
        }
        \caption{Reconstruction of randomly sampled images of digits 3 and 4 of Dynamic MNIST. (a) shows the original images of the reconstructed data. {\color{red}Red} bounding boxes show inaccurate reconstructions of digits with respect to their original images (e.g. some of the digit 4 images in ByPE-VAE could be misconstrued as digit 9). }
        %\vspace{-1.0em}
    \label{fig:RECONSTRUCTION1}
\end{figure*}

In Fig.~\ref{fig:RECONSTRUCTION1}, we compare the reconstruction of images in DMNIST for each method. The images are chosen through a random sampling of the latent space. Fig.~\ref{fig:RECONSTRUCTION1}(a) and (e) shows the original training images for the data being reconstructed, and (b)-(d), (f)-(h) show the reconstructions obtained by the different methods. % From the 10 runs for each method, we chose the model which resulted in the highest log likelihood on test data to reconstruct the images. 
%We highlight blurred reconstructions with red bounding boxes and incomplete or incorrect reconstructions with yellow bounding boxes. 
We observe that Hub-VAE reconstructed images are less blurred and have fewer inaccuracies compared to its competitors. 

\begin{figure*}[ht!]
 \centering
    \subfigure[Original Image ]{
        \centering
        \includegraphics[width=35mm,height=35mm]{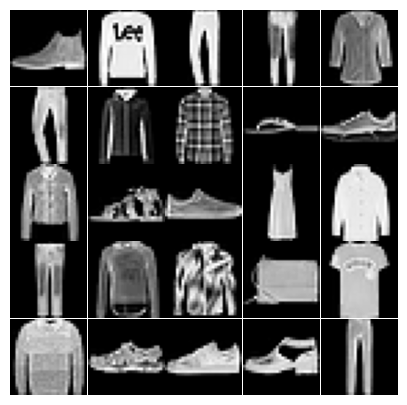}
        }
      \subfigure[Hub-VAE ]{
        \centering
        \includegraphics[width=35mm,height=35mm]{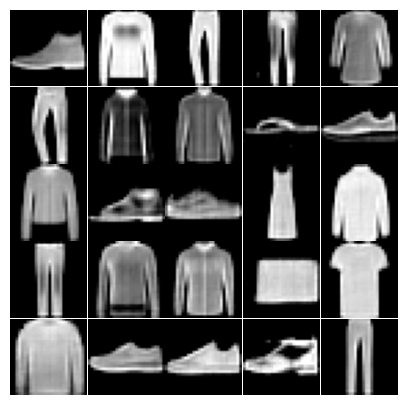}
        }
      \subfigure[Ex-VAE]{
        \centering
        \includegraphics[width=35mm,height=35mm]{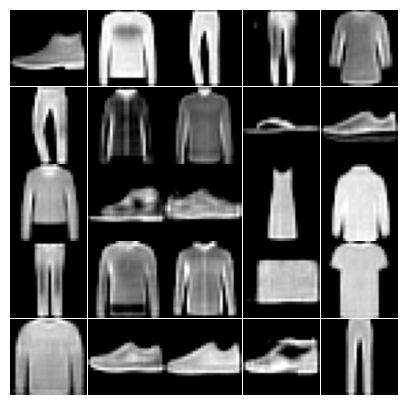}
        }
        \subfigure[ByPE-VAE]{
        \centering
        \includegraphics[width=35mm,height=35mm]{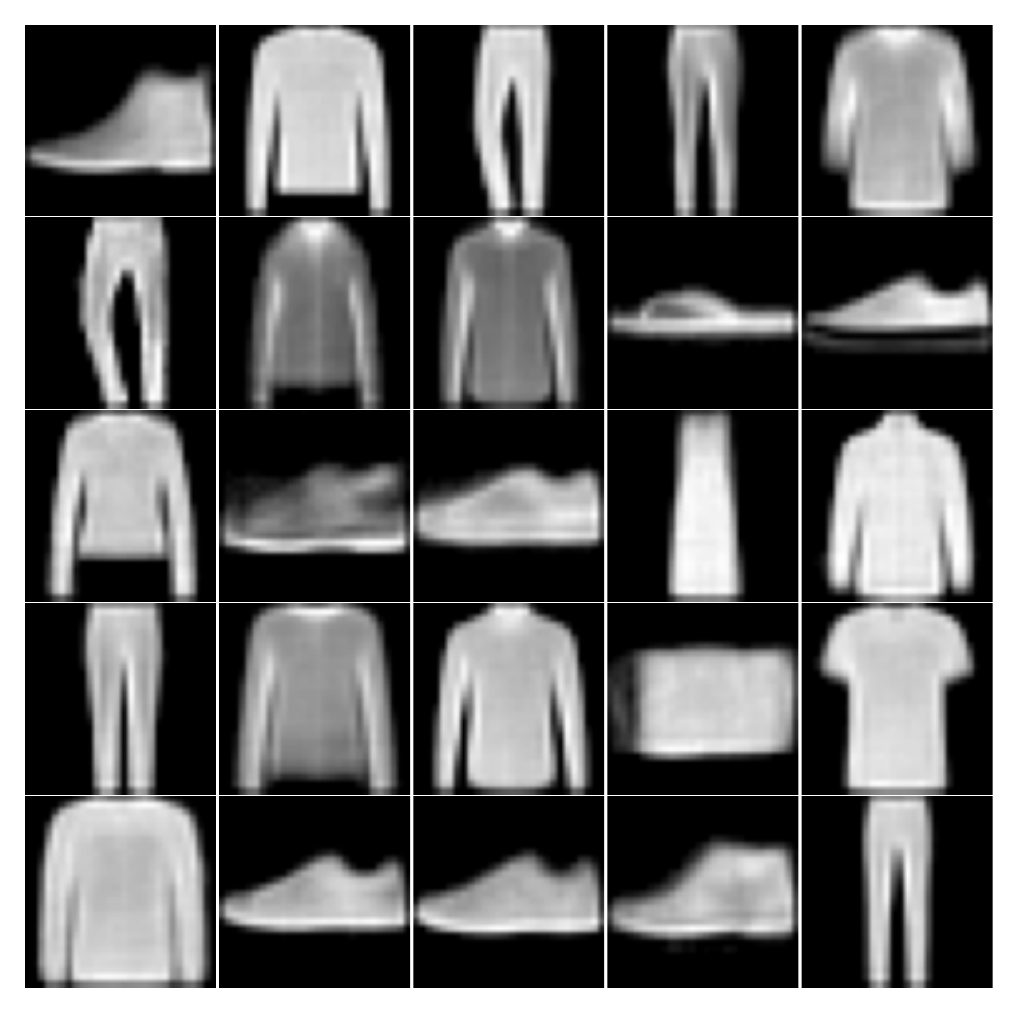}
        }
        
        \caption{Reconstruction of randomly sampled images of FMNIST from the latent space. (a) shows the original images of the reconstructed data.}
        %\vspace{-1.0em}
    \label{fig:RECONSTRUCTION2}
\end{figure*}

\begin{figure*}[ht!]
 \centering
    \subfigure[Original Image ]{
        \centering
        \includegraphics[width=35mm,height=35mm]{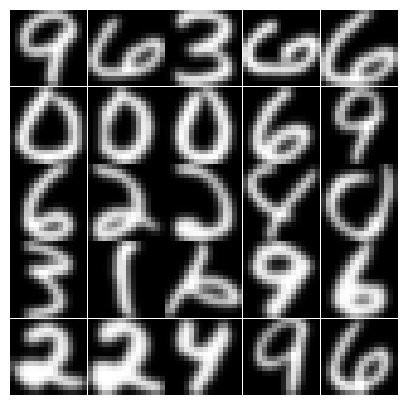}
        }
      \subfigure[Hub-VAE ]{
        \centering
        \includegraphics[width=35mm,height=35mm]{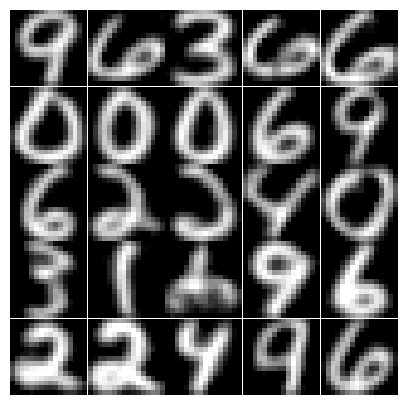}
        }
      \subfigure[Ex-VAE]{
        \centering
        \includegraphics[width=35mm,height=35mm]{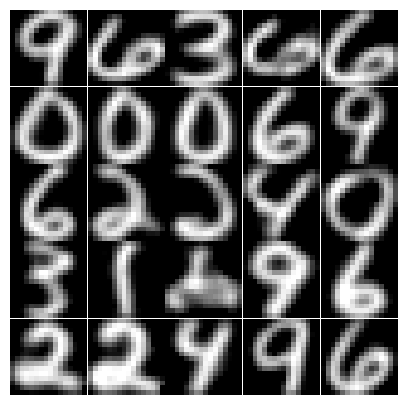}
        }
        \subfigure[ByPE-VAE]{
        \centering
        \includegraphics[width=35mm,height=35mm]{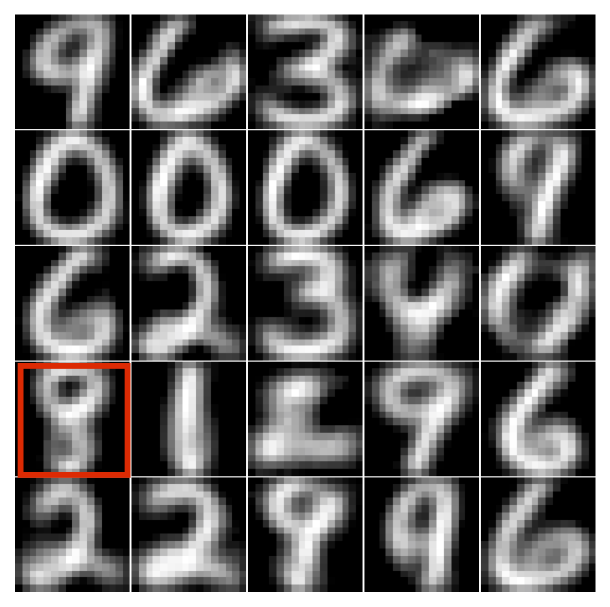}
        }
        
        \caption{Reconstruction of randomly sampled images of USPS from the latent space. (a) shows the original images of the reconstructed data. {\color{red} Red} bounding box on ByPE-VAE shows an incorrect reconstruction.}
        %\vspace{-1.0em}
    \label{fig:RECONSTRUCTION3}
\end{figure*}

The reconstructions for FMNIST in Fig.~\ref{fig:RECONSTRUCTION2} show an overall lesser variability among Hub-VAE and Ex-VAE. However, it can be observed that ByPE-VAE results in blurred images and fails to capture the finer details of the images which were captured by Hub-VAE and Ex-VAE (for example, class 'Sneaker', 'Ankle boots'). Similar results can be observed for USPS, where ByPE-VAE incorrectly reconstructs digit 8 instead of digit 3, as shown by red bounding box in Fig.~\ref{fig:RECONSTRUCTION3}. 

\begin{figure*}[ht!]
 \centering
      \subfigure[Hub-VAE ]{
        \centering
        \includegraphics[width=30mm,height=30mm]{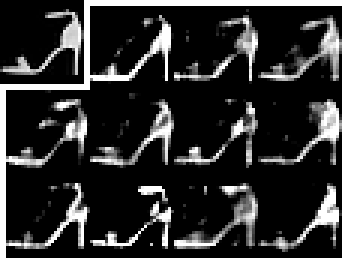}
        }
      \subfigure[Hub-VAE]{
        \centering
        \includegraphics[width=30mm,height=30mm]{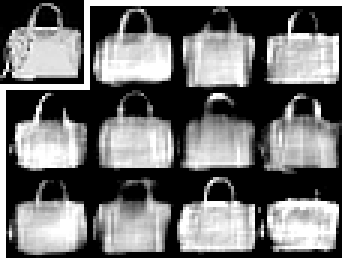}
        }
        \subfigure[Hub-VAE]{
        \centering
        \includegraphics[width=30mm,height=30mm]{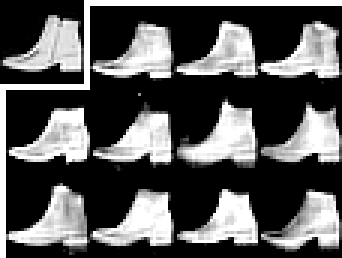}
        }
        \subfigure[Hub-VAE]{
        \centering
        \includegraphics[width=30mm,height=30mm]{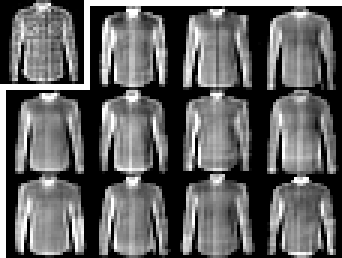}
        }\\
        
        \subfigure[Exemplar-VAE ]{
        \centering
        \includegraphics[width=30mm,height=30mm]{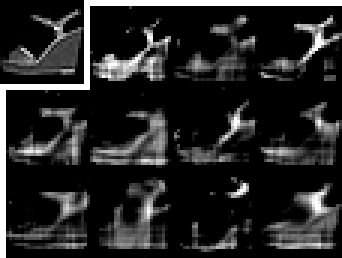}
        }
      \subfigure[Exemplar-VAE]{
        \centering
        \includegraphics[width=30mm,height=30mm]{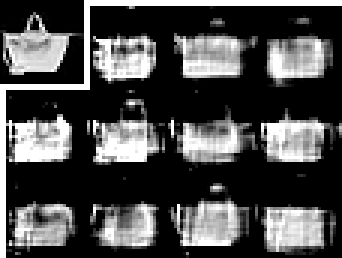}
        }
        \subfigure[Exemplar-VAE]{
        \centering
        \includegraphics[width=30mm,height=30mm]{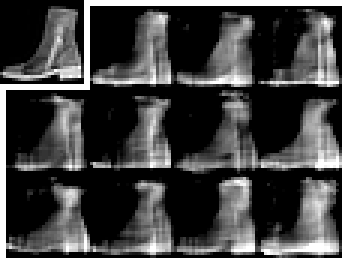}
        }
        \subfigure[Exemplar-VAE]{
        \centering
        \includegraphics[width=30mm,height=30mm]{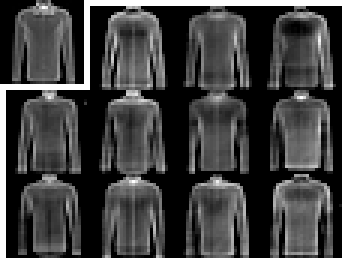}
        }
        \caption{Conditional image generation from chosen classes in FMNIST for (a)-(d) Hub-VAE and (e)-(f) Ex-VAE. The reference images (exemplar) are shown in the inset. }
        %\vspace{-1.0em}
    \label{fig:GENERATION2}
\end{figure*}

In  Fig.~\ref{fig:GENERATION2}, we compare the quality of conditional image generation on Hub-VAE and  Ex-VAE.% and ByPE-VAE.
% For Dynamic MNIST, we chose to generate images from the classes 3, 4, 8 and 9 as these classes exhibit features which can be confused with each other (3 and 8, 4 and 9), and they are embedded in close proximity in the latent space. It can be observed that the images generated by Hub-VAE are less blurred and closely follow the shape of the reference image (shown in the inset). Exemplar-VAE often generates blurred images and sometimes incorrect images (e.g., images generated for digit 9 do not accurately capture the reference image, and images resembling digit 8 are generated from digit 3). The above qualitative results show that a hub-based regularization of VAE is useful to improve the quality of the generative modeling. 
We plot the classes 'Sandals', 'Bag', 'Ankle Boots' and 'Shirt' for Fashion-MNIST. The classes 'Sandals' and 'Ankle Boots' can be confused with each other. We again observe the superior generative quality of Hub-VAE compared to Ex-VAE which generates blurred images and does not accurately capture the fine details of the reference image. 
%The generative ability of Exemplar-VAE may improve with a larger number of components in the prior distribution, as done in the experiments in \cite{EXEMPLAR-VAE} whose model uses $\ge$ 10000 components for Dynamic MNIST and Fashion MNIST. However, this also shows the strength of Hub-VAE which is able to achieve good quality results with a relatively smaller number of components, while Exemplar-VAE could not achieve a similar quality in generative modeling.

% % \begin{figure*}[ht!]
% %  \centering
% %     %   \subfigure[Hub-VAE ]{
% %     %     \centering
% %     %     \includegraphics[width=30mm,height=30mm]{pics/generated_3_hub2.png}
% %     %     }
% %       \subfigure[Hub-VAE]{
% %         \centering
% %         \includegraphics[width=30mm,height=30mm]{pics/generated_4_hub2.png}
% %         }
% %         % \subfigure[Hub-VAE]{
% %         % \centering
% %         % \includegraphics[width=30mm,height=30mm]{pics/generated_8_hub2.png}
% %         % }
% %         \subfigure[Exemplar-VAE]{
% %         \centering
% %         \includegraphics[width=30mm,height=30mm]{pics/generated_4_exm1.png}
% %         }
% %         \subfigure[Hub-VAE]{
% %         \centering
% %         \includegraphics[width=30mm,height=30mm]{pics/generated_9_hub2.png}
% %         }
% %         % \subfigure[Exemplar-VAE ]{
% %         % \centering
% %         % \includegraphics[width=30mm,height=30mm]{pics/generated_3_exm1.png}
% %         % }
% %         % \subfigure[Exemplar-VAE]{
% %         % \centering
% %         % \includegraphics[width=30mm,height=30mm]{pics/generated_8_exm1.png}
% %         % }
% %         \subfigure[Exemplar-VAE]{
% %         \centering
% %         \includegraphics[width=30mm,height=30mm]{pics/generated_9_exm1.png}
% %         }
% %         \caption{Conditional image generation from chosen classes in Dynamic MNIST and  Exemplar-VAE. The reference images (exemplar) are shown in the inset. }
% %         %\vspace{-1.0em}
% %     \label{fig:GENERATION1_A}
% \end{figure*}

\subsection{Running Time} We compare the running time in seconds of each method for an epoch in Table~\ref{tab:runtime}. VAE-Gaussian has the lowest running time. We observe that while Hub-VAE has a higher running time than the compared methods, the time is feasible from a practical standpoint. The bottleneck is the computation of hubs in each epoch, which involves pairwise distance computations. However, approximate nearest neighbor computation can be used to speedup the runtime of Hub-VAE.

\begin{table*}[ht!]
\begin{center}
\begin{tabular}{c|c|c|c|c|c}
\textbf{Data} & \textbf{VAE-Gaussian} & \textbf{VAE-Vamp} & \textbf{Ex-VAE} & \textbf{ByPE-VAE} & \textbf{Hub-VAE}\\
\hline
DMNIST & \bf{5.24} &	6.58 &	7.63 &	7.01 &	28.16 \\
FMNIST & \bf{5.46} &	6.64 &	11.12 &	7.65 &	33.62\\
USPS  & \bf{1.61} &	1.77 &	1.99 &	2.06 &	4.26 \\
Caltech101 & \bf{1.45} &	1.56 &	1.75 &	1.73 &	4.67 \\
CIFAR-10 & \bf{4.7}&	5.83 &	7.93 &	5.78 &	28.36 \\
%\hline
\end{tabular}
\end{center}
\caption{Running time (in seconds) per epoch.}
\label{tab:runtime}
\end{table*}

\subsection{Hub Characteristics}
\begin{figure*}[t]
 \centering
      
        \subfigure[USPS]{
        \centering
        \includegraphics[width=50mm,height=40mm]{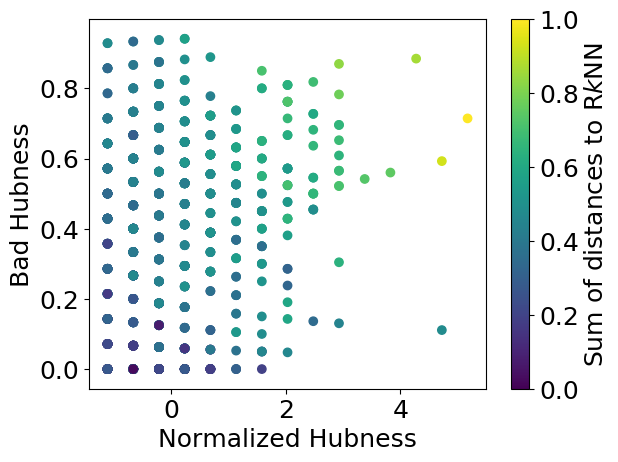}
        }
         \subfigure[Caltech101Silhouettes]{
        \centering
        \includegraphics[width=50mm,height=40mm]{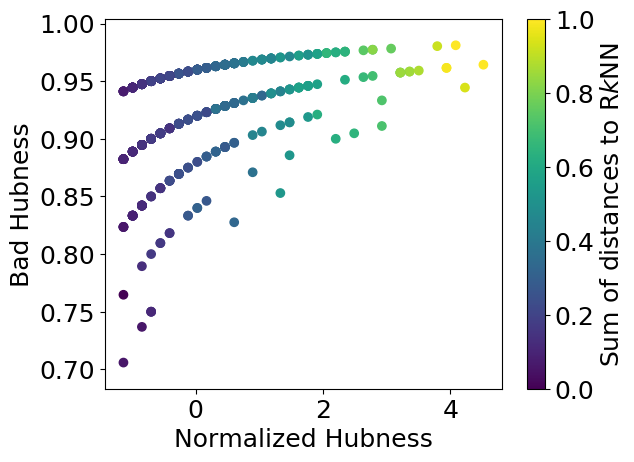}
        }
        
        \subfigure[USPS]{
        \centering
        \includegraphics[width=50mm,height=40mm]{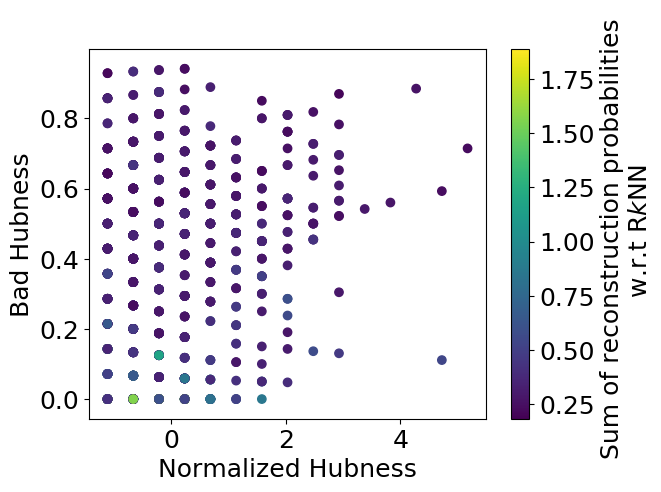}
        }
         \subfigure[Caltech101Silhouettes]{
        \centering
        \includegraphics[width=50mm,height=40mm]{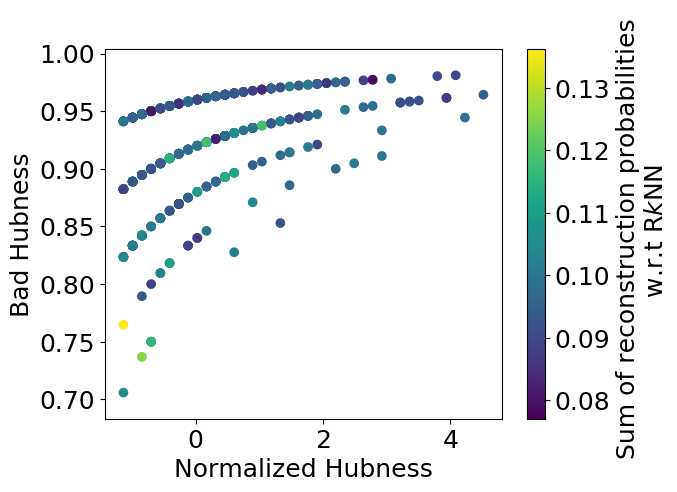}
        }
        
        \caption{ Scatter plots of characteristics of hubs.
        %pairwise 2-Wasserstein distances between hubs and their R$k$NNs in the latent space; (e)-(h) Scatter plots of reconstruction probabilities of hubs with respect to the distributions of their R$k$NNs.
        The $x$-axis in each sub-plot denotes normalized ($\mu_{N_k}$ = 0, $\sigma_{N_k}$ = 1) hubness scores. The $y$-axis denotes bad hubness.
        The hubs are color-coded by the sum of the pairwise distances to their reverse $k$-nearest neighbors (R$k$NN) in plots (a)-(b), and by their reconstruction probabilities w.r.t. the distributions of their R$k$NN in plots (c)-(d).
        Hubs with high pairwise distances (top-right quadrant of (a)-(b)) and low reconstruction probabilities  (bottom-left quadrant of (c)-(d)) with respect to their R$k$NN are strong bad hubs. %The distinction of strong bad hubs through the measured characteristics is more pronounced for DynamicMNIST, FashionMNIST, and Caltech101Silhouettes.
        % {\color{red}{Need to add labels of axes and fix current naming. Need to clarify how "bad hubness" is computed in this}}
        }
        %\vspace{-1.0em}
    \label{fig:hubc}
\end{figure*}

We show the hub characteristics for USPS and Caltech101 in Fig.~\ref{fig:hubc}
